\documentclass[letterpaper]{article}

\usepackage[T1]{fontenc}

\usepackage{geometry}
\usepackage{setspace}

\usepackage{achemso}
\setkeys{acs}{
  articletitle = true,   
  maxauthors   = 100     
}

\usepackage{graphicx}
\usepackage{float}
\newfloat{scheme}{htbp}{los}
\floatname{scheme}{Scheme}
\floatname{chart}{Chart}
\newfloat{graph}{htbp}{loh}

\usepackage{chemformula} 
\usepackage[version = 4]{mhchem} 

\usepackage{url}
\usepackage{subfigure}
\usepackage{multirow}
\usepackage{mhchem}
\usepackage{adjustbox}

\usepackage{multicol} 
\usepackage[utf8]{inputenc} 
\usepackage[T1]{fontenc}    
\usepackage{url}            
\usepackage{booktabs}       
\usepackage{amsfonts}       
\usepackage{bm}             
\usepackage{subcaption}     
\usepackage{array}
\usepackage{nicefrac}       
\usepackage{microtype}      
\usepackage{xcolor}         
\definecolor{tianyilan}{HTML}{66CCFF}
\usepackage{algorithm}
\usepackage{algpseudocode}
\usepackage{graphicx}
\usepackage{float} 
\usepackage{graphicx}
\usepackage{adjustbox}
\usepackage{trimclip}
\usepackage{varwidth}
\usepackage{booktabs}
\usepackage{multirow}
\usepackage[table,xcdraw]{xcolor}
\usepackage{subfigure}
\usepackage{enumitem}
\usepackage{amsmath,amssymb,amsthm}

\DeclareMathOperator*{\argmax}{arg\,max}
\usepackage{bm}

\usepackage{graphicx}
\usepackage{epstopdf}
\usepackage{hyperref}

\usepackage{authblk}
\author[1]{Mianzhi Liu{†}}
\author[2]{Fan Xiao{†}}
\author[2]{Zhiliang Yu}
\author[2]{Huayang Huang}
\author[2]{Yuke Li}
\author[3]{Yi Yang}
\author[4]{Wenbo Liu}
\author[2]{Yu Wu*}
\affil[1]{School of Cyber Science and Engineering, Wuhan University, Wuhan 430072, China}
\affil[2]{School of Computer Science, Wuhan University, Wuhan 430072, China}
\affil[3]{College of Computer Science and Technology, Zhejiang University, Hangzhou 310027, China}
\affil[4]{College of Chemistry and Molecular Sciences, Wuhan University, Wuhan 430072, China}

\title{RetroMPA: A Molecular Property-Aware Auxiliary Framework for Enhancing Retrosynthesis Prediction}
\date{*Email: wuyucs@whu.edu.cn}

\begin{document}

\maketitle

\begin{abstract}
  Retrosynthesis is a cornerstone of drug discovery and organic synthesis. While data-driven deep learning models have shown remarkable progress, they are designed to autonomously learn reaction patterns from extensive retrosynthesis datasets with limited explicit integration of established chemical knowledge as priors. 
  To address this limitation, we introduce \textbf{RetroMPA}, a molecular property-aware, post-hoc enhancement module that injects chemical knowledge into the retrosynthesis pipeline. Rather than functioning as an independent, standalone SMILES sequence generator from scratch, RetroMPA is conceptualized as a broadly applicable, model-agnostic chemical filter designed to recalibrate and optimize the predictive pathways of various existing algorithms. 
  This plug-and-play framework can be seamlessly integrated with a range of existing data-driven retrosynthesis methods, enhancing model outputs without necessitating any modifications to the original model architecture or requiring resource-intensive, model-specific retraining procedures. By operating at the molecular level and leveraging a property-aware latent embedding space, RetroMPA consistently improves top-1 accuracy across eight representative retrosynthesis models by an average of \textbf{5.50}\% on USPTO-50K. 
  Furthermore, we demonstrate its scalability by validating its performance on the large-scale USPTO-Full dataset, achieving an average improvement of about \textbf{2.03}\% across both template-based and template-free architectures. 
  In addition, wet-lab experiments provide preliminary support for the practical utility of the framework. 
  These syntheses confirmed viable, previously unreported substrate combinations for established, classic reaction paradigms—specifically, the Suzuki–Miyaura coupling, the Bucherer reaction, and the Friedel–Crafts acylation, thereby suggesting that RetroMPA can operate beyond mere data fitting. The code is open-sourced at \href{https://github.com/MengzhouLu/RetroMPA}{https://github.com/MengzhouLu/RetroMPA}.
\end{abstract}

\section*{Keywords}

Retrosynthesis Prediction, Molecular Property Awareness, Knowledge-Constrained Learning, Molecular Representation Learning, Chemical Semantic Space.




\begin{figure}[t] 
\captionsetup{justification=justified, singlelinecheck=false}
\centering 
\includegraphics[width=1\textwidth,trim=2.5cm 0.5cm 3.1cm 0.88cm, clip]{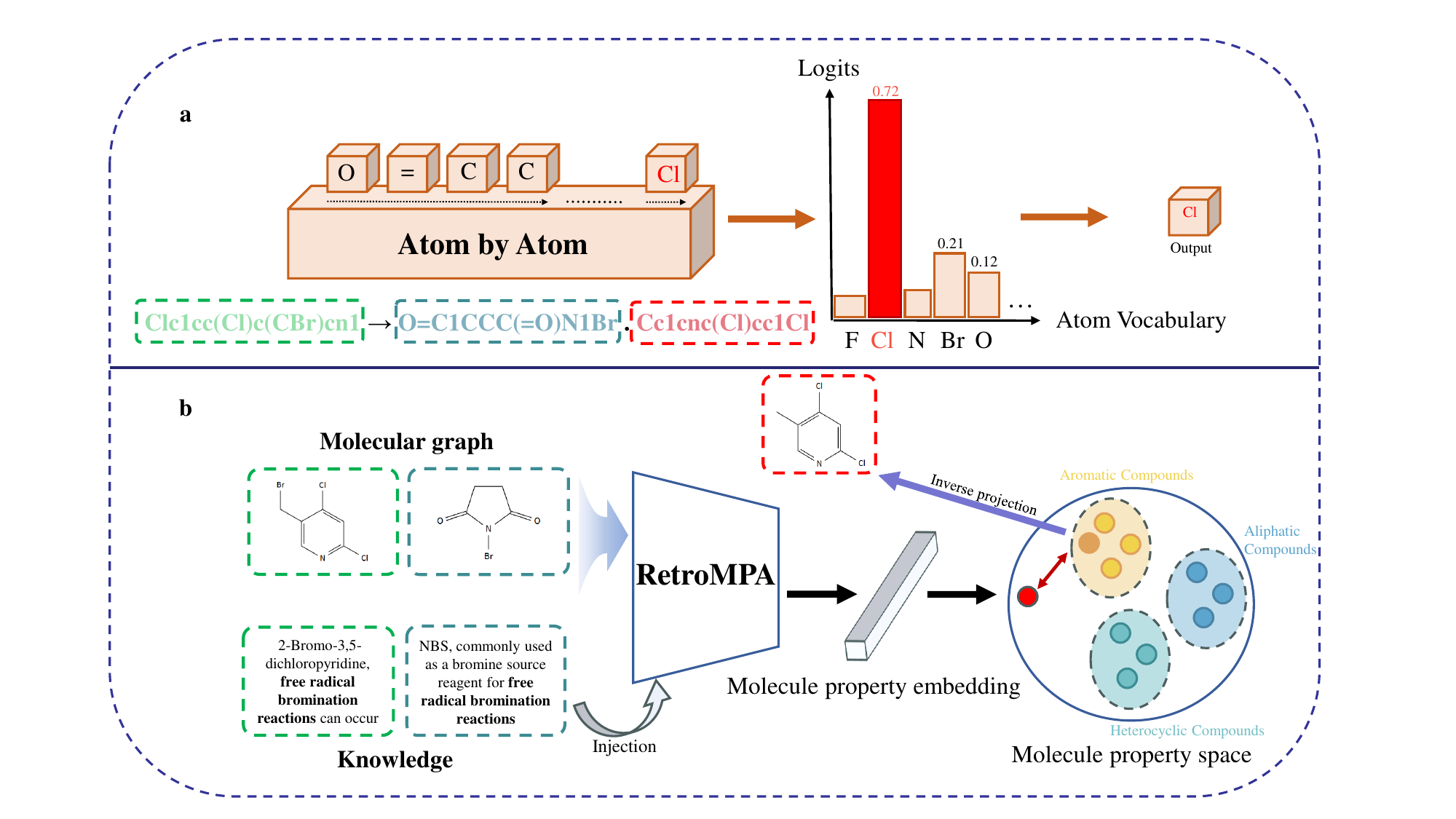} 

\caption{\textbf{Standard retrosynthesis autoregressive modeling (Existing) \textit{vs} our proposed RetroMPA (Ours).}
(a) Autoregressive applied to SMILES: sequential atom token generation from left to right, atom by atom, with limited explicit integration of chemical knowledge as priors; (b) RetroMPA generates molecular property embeddings guided by chemical knowledge, subsequently outputting complete molecules through inverse projection in chemically constrained spaces.  
} 
\label{fig:motivation} 
\end{figure}

\section{Introduction}

Retrosynthesis~\cite{long2025artificial, torren2024models, li2026retro} is the process of deducing feasible synthetic routes for a target molecule by recursively decomposing it into simpler precursors.          At the heart of this process lies single-step retrosynthesis, which identifies plausible reactants for a given product~\cite{maziarz2025re}.          While decades of chemical intuition and empirical knowledge have guided this task, the explosive growth of reaction data has spurred the development of data-driven computational tools to assist or automate retrosynthetic planning~\cite{somnath2021learning}.

Current computational retrosynthesis methods commonly employ SMILES strings and molecular graphs for representation. SMILES strings \cite{weininger1988smiles} compress complex non-linear molecular structures into linear sequences. 
However, its syntax proves empirically difficult to learn with standard sequence models, requiring complex model designs and massive data to overcome the grammatical dependencies of linear representations \cite{wang2021chemical},  while graphs naturally preserve structural features \cite{li2023knowledge,doi:10.1021/acs.jcim.9b00538}. Molecular representation learning  aims to map molecules to latent vectors that encode structural-property relationships. SMILES-based methods such as \textit{SMILES-BERT} \cite{wang2019smiles} leverage masked language modeling, while the graph-based approach \textit{MolR} \cite{wang2021chemical} preserves reaction equivalence through GNNs.

Concurrently, physics-based simulations and emerging innovations in physics- and knowledge-based approaches play a crucial role in modern computational chemistry, particularly within the domains of physics- and quantum-chemistry-informed strategies for retrosynthesis and synthesis planning. Sumiya et al. developed a quantum chemical approach for tracing reaction paths backward from a given target compound and enumerating possible reactant candidates, demonstrating that first-principles reaction-path exploration can provide mechanistically grounded hypotheses for reactant prediction\cite{doi:10.1021/jacsau.2c00157}. Toniato et al. further examined the use of on-demand quantum chemical data generation to support machine-learning reaction and retrosynthesis planning, especially for low-confidence predictions where additional first-principles information may help improve decision confidence or provide data for subsequent model refinement\cite{D3DD00006K}. In addition, Liu et al. proposed a reaction-kinetics-based retrosynthesis planning framework using a transition-state automated generation method, illustrating how kinetic feasibility can be incorporated into the evaluation of candidate synthetic pathways\cite{liu2023article}. Physics-based and quantum-chemical approaches can offer mechanistic, energetic, or kinetic information that is difficult to obtain from purely data-driven models, but they are typically computationally more demanding because they may require explicit reaction-path exploration, transition-state generation, or first-principles calculations, which can limit their direct use in large-scale global retrosynthetic pathway planning.

In recent years, deep learning has emerged as a transformative force in retrosynthesis planning~\cite{zhong2024recent, guo2025directly}, with approaches broadly categorized into template-based, semi-template-based, and template-free methods.           Template-based methods rely on predefined reaction rules to ensure chemical validity, but their applicability is inherently limited to predefined templates~\cite{hartenfeller2011collection,szymkuc2016computer,coley2017prediction,designer2009retrosynthetic}.           Semi-template-based methods offer greater flexibility by identifying reaction centers via atom mapping and generating synthons~\cite{yao2024node,somnath2021learning,yan2020retroxpert}, but often depend on atom-mapping information unavailable in real-world prediction scenarios\cite{maziarz2025re}.           
Template-free models, which treat retrosynthesis as a sequence translation task~\cite{han2024retrosynthesis,jiang2023learning,zhong2022root,wan2022retroformer,Sacha2021Graph}, have achieved state-of-the-art performance recently by leveraging large reaction corpora.

Despite these successes, existing retrosynthesis approaches ~\cite{zhong2024recent, guo2025directly} concentrate on designing sophisticated architectures to learn more reaction rules primarily from reaction data statistics, which still lack sufficient exploration in leveraging molecular chemical properties.     Contemporary models are predominantly data-driven, learning statistical patterns of atom arrangements with less emphasis on explicit chemical reasoning mechanisms.       Such models operate at the atom or token level, generating SMILES strings~\cite{weininger1988smiles} autoregressively without holistic molecular understanding, leading to chemically implausible outputs (Fig.~\ref{fig:motivation}a).           This arises from two fundamental gaps: (1) Overreliance on data-driven pattern learning without reaction-relevant molecular property knowledge as chemical priors, (2) unconstrained model outputs lacking chemical knowledge constraints.

Chemical reactions originate from molecular interactions, where molecular properties help determine reaction feasibility, pathways, and efficiency~\cite{hammond1955correlation, kolb2001click, noyori2002asymmetric}. Therefore, integrating molecular-property knowledge into retrosynthesis models is a reasonable and potentially valuable direction. Chemists often infer possible reactions and reactants by analyzing relationships between the chemical properties of products and reactants~\cite{corey1967general}. Thus, integrating molecular property awareness into retrosynthesis models may provide more than an incremental improvement and could contribute to more chemically plausible and generalizable predictions.

However, existing retrosynthesis base models have undergone lengthy and expensive training or optimization processes. Requiring these foundation models to be restructured or retrained in order to incorporate chemical knowledge would incur substantial computational overhead. A more efficient and cost-effective approach is needed to inject knowledge into already well-performing models and improve their performance.

In this work, we introduce RetroMPA, a novel molecular property-aware auxiliary framework designed to enhance existing retrosynthesis models without architectural modification or retraining.           
Our approach is inspired by the chemical intuition that reactions can be understood as transformations in property space rather than mere rearrangements of atoms.           
RetroMPA first learns molecular chemical properties through multimodal knowledge of molecular-chemical text. Subsequently, RetroMPA utilizes diverse reactions to leverage chemical properties for retrosynthesis by learning latent correlations between product and reactants based on their properties. Departing from conventional methods, our method reduces the dependency on reaction templates or atom vocabularies. Instead, its prediction mechanism involves nearest-neighbor retrieval in a molecular embedding space, followed by inverse projection to generate reactants(Fig.~\ref{fig:motivation}b). Finally, the framework further applies chemically grounded constraints to refine the output of the base model. Importantly, RetroMPA itself does not rely on predefined reaction templates or reaction class labels. It imposes no requirements on the base model and is plug-and-play—requiring no modifications to the model architecture or retraining.
RetroMPA is structurally positioned as a lightweight, complementary algorithmic tool rather than a direct replacement for foundational physical simulations. By systematically injecting chemical priors into the model, our framework is designed to provide a rapid, preliminary chemical constraint filter.

We first introduce the RetroMPA framework briefly, which enhances existing retrosynthesis predictors by integrating molecular property knowledge as chemical priors, enabling more chemically plausible and interpretable predictions.   We then assess its impact on the standard USPTO-50K benchmark~\cite{schneider2016s} across eight diverse base models, showing consistent improvements in top-1 accuracy without retraining.        
On the more complex USPTO-Full dataset, RetroMPA likewise demonstrates performance improvements, indicating its generalizability.
Next, we evaluate the effectiveness and computational efficiency of the dynamic molecular dictionary under varying retrieval sizes and vocabulary scales.           Through case studies, we further illustrate how RetroMPA leverages chemical semantics to guide reasoning.           We further validate the model’s generalization capability through wet-lab experiments on previously unreported substrate combinations within established reaction classes, including Suzuki–Miyaura coupling, the Bucherer reaction, and Friedel–Crafts acylation.            Together, these evaluations provide a comprehensive view of how RetroMPA improves prediction accuracy, chemical plausibility, and model-agnostic adaptability in retrosynthesis, while also demonstrating its potential for discovering synthetically viable routes beyond training data.      A more elaborate description of model architectures and rationale for certain design choices is given at the end of the paper.

\begin{figure}[t] 
\centering 
\includegraphics[width=1\textwidth,trim=4cm 2.3cm 4cm 1cm, clip]{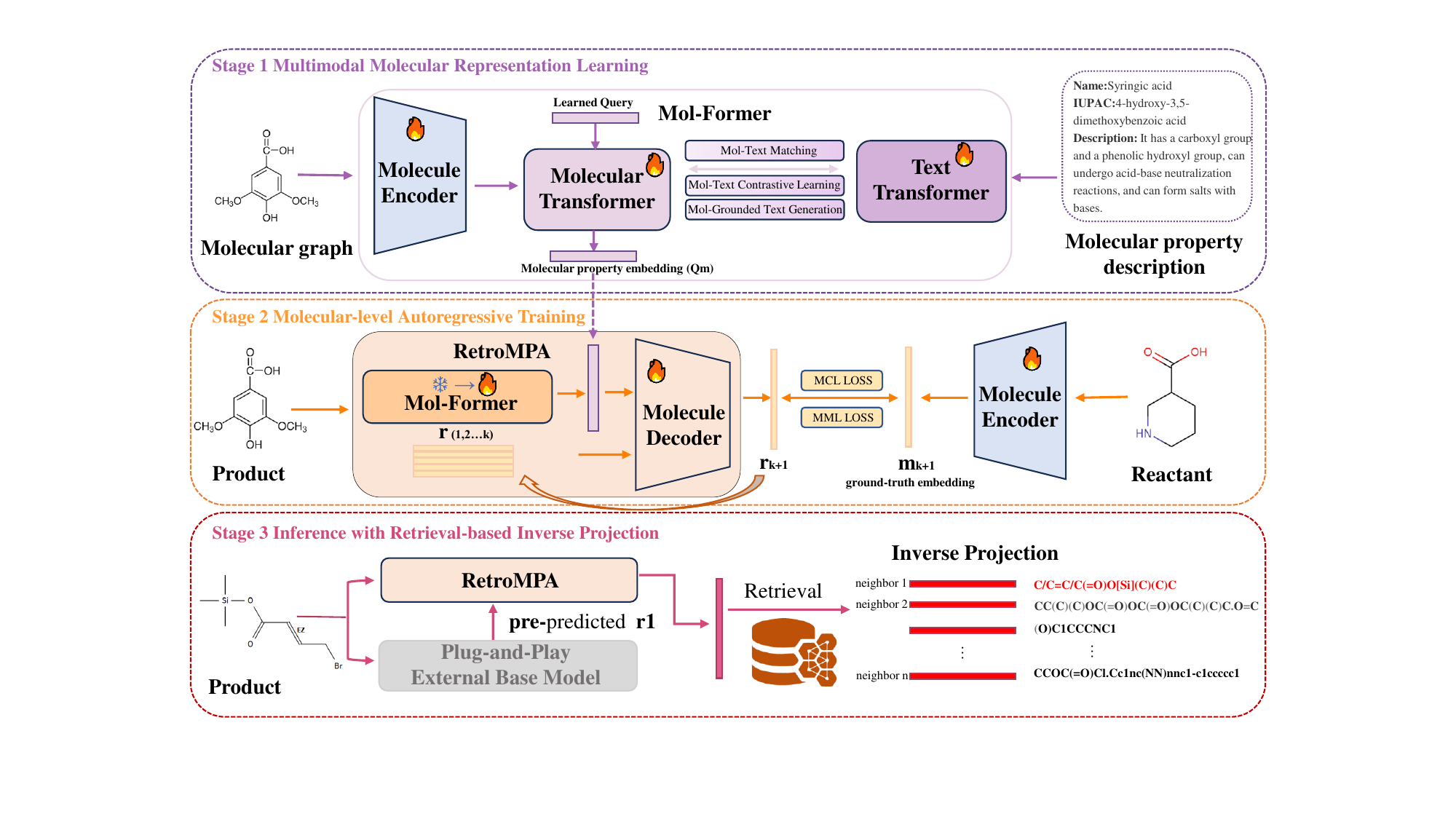} 
\caption{\textbf{Overview of the three-stage RetroMPA framework. Stage 1: }Mol-Former encodes molecules into molecular property embeddings $\mathbf{Q}_m$ through multimodal molecule–text pretraining. \textbf{Stage 2: }Given the product embedding \(Q_p\) and previously available or predicted reactants \( (r_1, r_2, \dots, r_{k}) \), the Molecular Decoder predicts the embedding of the next reactant \( r_{k+1} \). \textbf{Stage 3:} 
The predicted embedding is mapped to molecular SMILES by Inverse Projection.}
\label{fig:method} 
\end{figure}

\section{Results and discussion}
\label{Results and discussion}
\subsection{Overview of the RetroMPA Framework}

RetroMPA is a knowledge-guided, plug-and-play auxiliary framework designed to enhance the chemical plausibility of single-step retrosynthesis predictions generated by any existing data-driven model.              The framework operates in three synergistic stages, as illustrated in Fig.~\ref{fig:method}: \textbf{Stage 1:} Mol-Former encodes molecules into molecular property embeddings, trained on multimodal molecular-chemical text data.     A multimodal encoder, Mol‑Former, aligns molecular structures with textual chemical descriptions to generate molecular property embeddings that encode functional groups, stereochemistry, and other chemically meaningful attributes. \textbf{Stage 2:} Molecular‑Level Prediction.      A molecular-level decoder, termed Molecular Decoder, learns the transformation patterns of molecular properties from products to reactants. Conditioned on the product's molecular property embedding, it infers the desired molecular properties of reactants and generates the embeddings of the expected reactants.         
This molecular‑level generation avoids the combinatorial complexity of atom‑by‑atom decoding. \textbf{Stage 3:} Output molecular SMILES through Inverse Projection.   Decoding the predicted reactant embeddings into concrete molecular SMILES by performing nearest-neighbor search against a dynamic molecular dictionary.       The dictionary can be updated with molecules from base‑model predictions or real‑world chemical catalogs, enabling adaptation to novel chemical spaces.

RetroMPA operates post hoc on the raw reactant predictions of a base model, refining them by incorporating chemical knowledge through the explicit integration of molecular property semantics as chemical priors.
As a simple example, given a product $P$, the base model predicts a reactant pair $(b_1, b_2)$. RetroMPA first encodes the product $P$ and a reactant from $(b_1, b_2)$ into vectors (Stage 1), then predicts the vector of the remaining reactant (Stage 2), and finally projects the predicted vector back to a molecule (Stage 3). This process yields a complete, chemically constrained reactant set, thereby refining the base model’s prediction.
The core insight is that chemical reactions are governed by the intrinsic properties of participating molecules, not merely by statistical co-occurrence in reaction datasets.               By constraining predictions within a chemically meaningful embedding space, RetroMPA improves both the reliability and generalizability of retrosynthetic results.               Full technical details of the framework are provided in Section 4.              Here, we focus on its chemical utility and empirical impact.

\subsection{Performance Enhancement on the USPTO‑50K Benchmark}
To assess the compatibility and general utility of RetroMPA within established computational retrosynthesis workflows, we evaluated its impact on the widely adopted USPTO‑50K benchmark~\cite{schneider2016s}.            Following prior work, we report top-1 accuracy based on exact match of canonical SMILES between prediction and ground truth, without using atom mappings or reaction class labels, reflecting real-world deployment conditions.

RetroMPA was integrated post-hoc with eight representative base models spanning major retrosynthesis paradigms: template-based (LocalRetro~\cite{chen2021deep}), semi-template-based (GraphRetro~\cite{somnath2021learning}) and template-free (Transformer\cite{karpov2019transformer}, Retroformer~\cite{wan2022retroformer}, R‑SMILES~\cite{zhong2022root}, NAG2G~\cite{yao2024node}, EditRetro~\cite{han2024retrosynthesis}, RetroWise~\cite{zhang2024retrosynthesis}).              Crucially, no retraining, fine-tuning, or architectural modification of the base models was performed;              RetroMPA operated in a plug-and-play fashion using only the initial prediction of the base model and the structure of the product as input. To ensure fair comparison, all baseline results are re-evaluated using publicly released model weights.

As summarized in Table~\ref{tab:usp50k}, RetroMPA consistently improved top-1 accuracy across all eight models, with average gains of 5.50\% without retraining or fine-tuning the base models.                  
The gains spanned all methodological categories, demonstrating the framework’s paradigm‑agnostic applicability. The Transformer model, which lacks explicit chemical inductive biases, exhibited the highest absolute improvement (+8.19\%).                
This suggests that RetroMPA effectively compensates for the absence of chemical priors in purely data‑driven sequence‑to‑sequence architectures by injecting property‑aware constraints. 
Both LocalRetro and GraphRetro, which rely on reaction templates or centers, showed improvements exceeding 5\%.               
This indicates that RetroMPA complements rule‑based approaches by enhancing their ability to handle chemically plausible but template‑ambiguous cases. Notably, even models already achieving high baseline performance, such as EditRetro (60.34\% → 64.75\%) and RetroWise (64.71\% → 68.30\%)—benefited from RetroMPA’s chemical priors.
             These improvements are particularly meaningful given the already high baseline performance and competitive benchmark setting.

These results collectively demonstrate that injecting molecular property knowledge as an auxiliary constraint can consistently enhance retrosynthesis prediction across diverse methodological paradigms.           The fact that RetroMPA operates in a plug‑and‑play fashion further underscores its practical utility: existing deployed models can be seamlessly upgraded without costly retraining or architectural overhaul.                   It does not rely on specific internal representations or training objectives of the base model;        instead, it leverages only the molecular property semantics embedded in the reactant–product pair.                      This universality makes RetroMPA a practical and low-overhead enhancement for existing retrosynthesis pipelines, particularly in industrial settings where model retraining is costly or infeasible.

In summary, while the USPTO-50K benchmark serves primarily as a compatibility testbed, the consistent performance gains observed across heterogeneous models suggest that integrating molecular property knowledge as a chemical prior can improve the reliability and chemical plausibility of retrosynthetic predictions.

\begin{table}[t]
\centering
\caption{Top-1 accuracy on USPTO-50K dataset. $\Delta$ indicates the improvement after using RetroMPA. Results of our method are averaged over 3 random seeds, reported as mean $\pm$ standard deviation.}
\label{tab:usp50k}
\begin{tabular}{@{}lccc@{}}
\toprule
\textbf{Method} & \textbf{Original Acc.} & \textbf{+Ours} & $\Delta$ \\ \midrule
\multicolumn{4}{l}{\textit{Template-based}} \\
\quad LocalRetro \cite{chen2021deep} & 52.59 & 57.76 $\pm$ 0.06 & \textbf{5.17} $\pm$ 0.06 \\ 
\addlinespace
\multicolumn{4}{l}{\textit{Semi-template-based}} \\
\quad GraphRetro \cite{somnath2021learning} & 53.72 & 59.54 $\pm$ 0.12 & \textbf{5.82} $\pm$ 0.12 \\ 
\addlinespace
\multicolumn{4}{l}{\textit{Template-free}} \\
\quad Transformer \cite{karpov2019transformer} & 42.80 & 50.99 $\pm$ 0.16 & \textbf{8.19} $\pm$ 0.16 \\
\quad Retroformer \cite{wan2022retroformer} & 52.73 & 58.68 $\pm$ 0.08 & \textbf{5.95} $\pm$ 0.08 \\
\quad R-SMILES \cite{zhong2022root} & 53.07 & 58.40 $\pm$ 0.10 & \textbf{5.33} $\pm$ 0.10 \\
\quad NAG2G \cite{yao2024node} & 54.66 & 60.20 $\pm$ 0.14 & \textbf{5.54} $\pm$ 0.14 \\
\quad EditRetro \cite{han2024retrosynthesis} & 60.34 & 64.75 $\pm$ 0.08 & \textbf{4.41} $\pm$ 0.08 \\
\quad RetroWise \cite{zhang2024retrosynthesis} & 64.71 & 68.30 $\pm$ 0.06 & \textbf{3.59} $\pm$ 0.06 \\ \bottomrule
\end{tabular}
\end{table}

\begin{table}[t]
\centering
\caption{Top-1 accuracy on USPTO-Full dataset. $\Delta$ indicates the improvement after using RetroMPA. Results of our method are averaged over 3 random seeds, reported as mean $\pm$ standard deviation.}
\label{tab:uspfull}
\begin{tabular}{@{}lccc@{}}
\toprule
\textbf{Method} & \textbf{Original Acc.} & \textbf{+Ours} & $\Delta$ \\ \midrule
\multicolumn{4}{l}{\textit{Template-based}} \\
\quad LocalRetro \cite{chen2021deep} & 40.12 & 42.92 $\pm$ 0.04 & \textbf{2.80} $\pm$ 0.04 \\ 
\addlinespace
\multicolumn{4}{l}{\textit{Template-free}} \\
\quad EditRetro \cite{han2024retrosynthesis} & 51.10 & 52.36 $\pm$ 0.03 & \textbf{1.26} $\pm$ 0.03 \\ \bottomrule
\end{tabular}
\end{table}

\subsection{Performance Enhancement on the USPTO‑Full Benchmark}

While performance on highly curated benchmarks like USPTO-50K confirms basic algorithmic validity, a rigorous evaluation of the framework's broad applicability necessitates stress-testing against larger-scale datasets. To this end, we expanded our benchmarking protocols to encompass the diverse USPTO-Full dataset. 
Due to computational constraints and the restricted availability of open-source weights for certain base models on USPTO-Full, our scalability evaluation focused strictly on two representative paradigms: LocalRetro (template-based graph models) and EditRetro (template-free string manipulation mechanisms).

The empirical results summarized in Table~\ref{tab:uspfull} suggest that RetroMPA remains effective under a larger and more diverse data setting. The application of our post-hoc auxiliary module yielded consistent improvements across the two evaluated paradigms, delivering a 2.80\% absolute gain to the template-based LocalRetro and a 1.26\% improvement to the highly optimized EditRetro. This observation corroborates that RetroMPA operates effectively as a scalable, universal inference layer, preserving its capacity for property-aware chemical reasoning under the evaluated USPTO-Full setting.

\subsection{Relationship between Base Model Performance and RetroMPA Improvement}
We further analyzed the correlation between the intrinsic performance of base models and the magnitude of improvement delivered by RetroMPA. The analysis suggests a compensatory trend between baseline performance and the magnitude of RetroMPA-induced improvement. 
The improvement brought by RetroMPA is related to the performance of the base model. As the base model attains higher accuracy, fewer incorrect predictions remain available for correction, and accordingly, the improvement conferred by RetroMPA generally diminishes. 

\begin{figure}[t]
    \centering
    \includegraphics[width=0.8\textwidth]{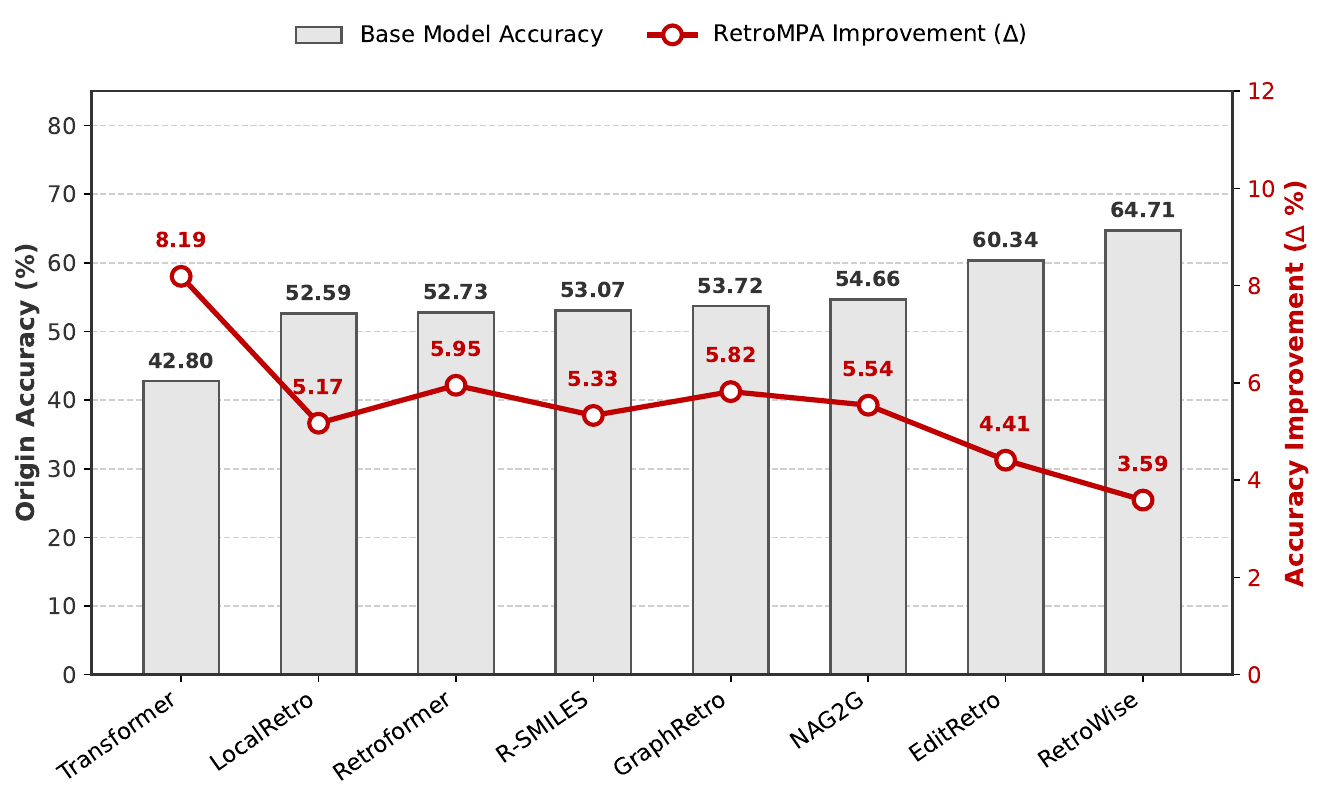}
    \caption{\textbf{Performance enhancement achieved by RetroMPA across various base models on USPTO-50K.} The gray bars indicate the original top-1 accuracy of the base models (left $y$-axis), arranged in ascending order of their inherent capabilities. The red line illustrates the absolute accuracy improvement ($\Delta$) conferred by RetroMPA (right $y$-axis).} 
    \label{fig:performance_gain_combo}
\end{figure}

\begin{figure}[t]
    \centering
    \includegraphics[width=0.65\textwidth]{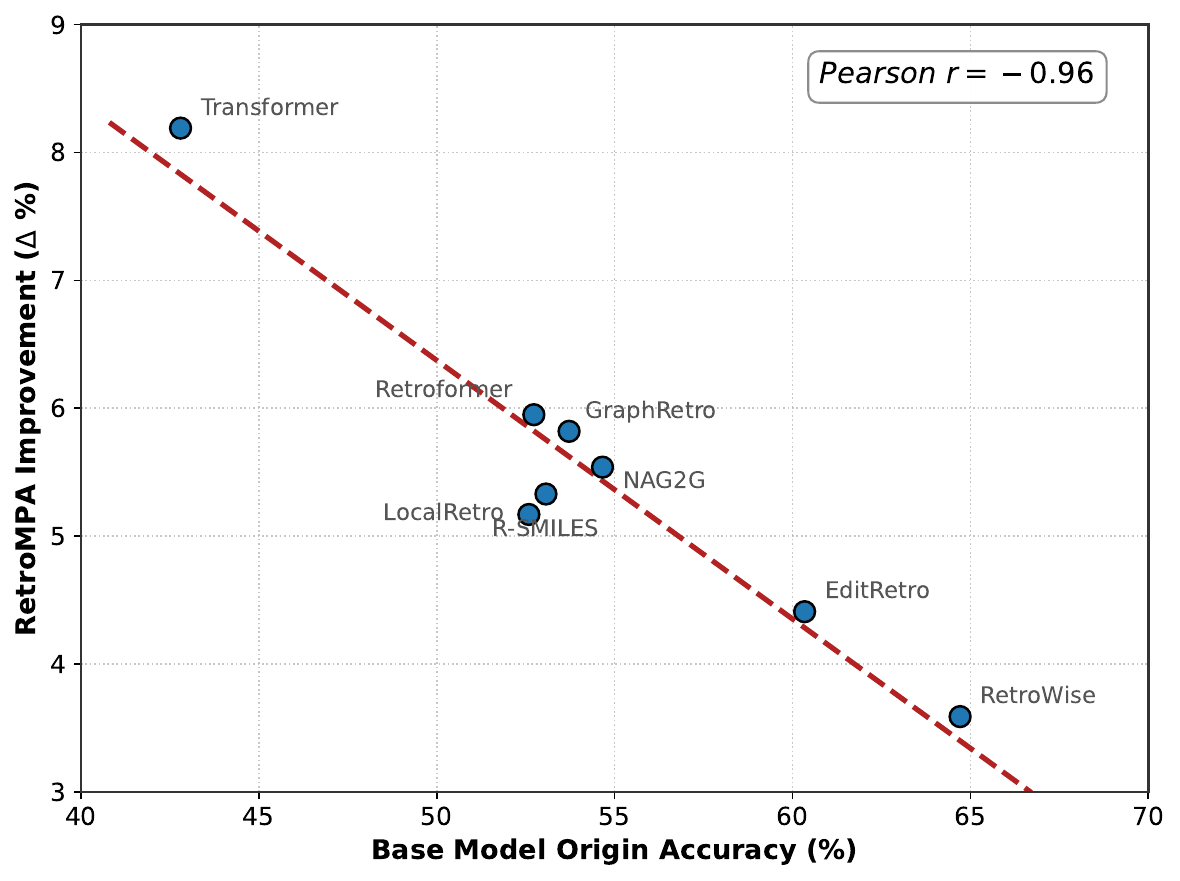}
    \caption{\textbf{Correlation analysis between the performance of base models and the magnitude of improvement delivered by RetroMPA on USPTO-50K.} The negative Pearson correlation ($r = -0.96$) shows a strong negative association.}
    \label{fig:performance_correlation}
\end{figure}

The overall trend observed in benchmark evaluations (see Figure \ref{fig:performance_gain_combo} and Figure \ref{fig:performance_correlation}) indicates that the weaker the base model, the greater the absolute improvement conferred by RetroMPA. When applied to the weakest baseline models (e.g., a sequence-to-sequence Transformer lacking chemical knowledge), RetroMPA yields the largest absolute improvements. Conversely, as the base model’s inherent capability strengthens and approaches advanced performance (e.g., EditRetro), the absolute performance gain provided by RetroMPA diminishes. Nevertheless, this improvement remains appreciable in magnitude. This suggests that RetroMPA serves as a corrective aid for fundamental errors in weaker models, while functioning as a chemistry-informed constraint filter for advanced models.

This trend was also observed on the more complex and chemically diverse USPTO-Full dataset. The substantially increased prediction difficulty reduced both the performance of the base models and the corresponding improvement magnitude; however, this further underscores the appreciable value of the gains brought by RetroMPA. It retained the capacity to capture and exploit molecular-property interactions in retrosynthesis within complex reaction relationships, without compromising compatibility with existing base models.

\subsection{Effectiveness and Efficiency of RetroMPA}

RetroMPA ultimately retrieves molecules from a molecular dictionary and outputs reactants, thereby bridging the continuous embedding space generated by the molecular decoder to discrete, chemically valid molecular structures.       Therefore, it is essential to further investigate the generation effectiveness and retrieval efficiency of RetroMPA beyond its performance on benchmark datasets.

To assess practical applicability, we report the computational overhead and latency introduced by RetroMPA.
The universal pretraining phase of the Mol-Former and the optimization of the Molecular Decoder are resource-intensive. Leveraging a high-performance computing cluster equipped with four NVIDIA RTX 4090 GPUs and driven by dual Intel Xeon Platinum 8352V CPUs, the comprehensive network training demanded approximately 130 hours. However, it should be emphasized that this training regimen represents a one-time amortized sunk cost. Because RetroMPA learns the universal relationships and patterns of chemical property changes between products and reactants, this singular pretrained artifact can be subsequently attached to \textit{any} base model without requiring a single epoch of fine-tuning or secondary backpropagation.
Crucially, the inference overhead introduced during active prediction is small. RetroMPA requires approximately 0.014 seconds, or 14 ms, per single inference. For context, reported single-sample inference latencies for existing retrosynthesis models are 177.4 ms for EditRetro and 292.1 ms for R-SMILES, excluding model loading time. Therefore, the additional per-inference overhead introduced by RetroMPA is substantially smaller than these reported inference times, approximately 12.7× lower than EditRetro and 20.9× lower than R-SMILES. Considering that RetroMPA is a plug-and-play framework that delivers improvements across the tested base models, these results suggest its potential practical utility for industrial applications.

\begin{figure}[t] 
\centering 
\includegraphics[width=1\textwidth,trim=0cm 15cm 0cm 0cm, clip]{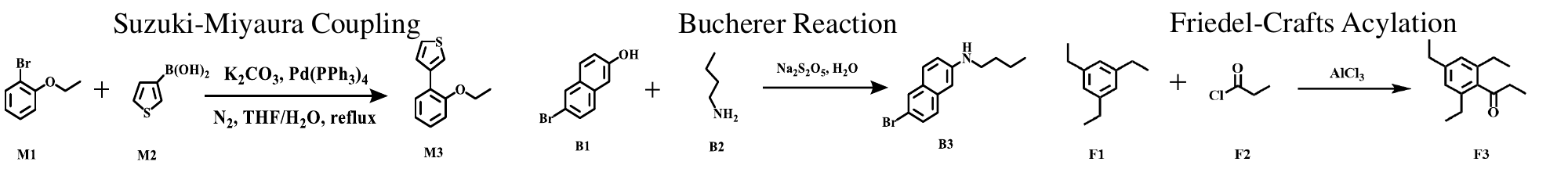} 
\caption{\textbf{Experimentally Validated Previously Unreported Substrate Combinations Predicted by RetroMPA Using Dynamic Molecular Dictionary.}
RetroMPA, leveraging its dynamic molecular dictionary, predicted three unreported substrate combinations, including Suzuki–Miyaura coupling, Bucherer reaction, and Friedel–Crafts acylation.
} 
\label{fig:wet} 
\end{figure}

An important requirement for retrosynthesis tools intended for real-world synthetic planning is their ability to generalize beyond known reaction databases and propose viable routes for novel molecules.
To this end, RetroMPA employs a dynamic molecular dictionary—a key component that enables adaptation to unseen chemical space by incrementally incorporating candidate molecules (e.g., from base model predictions or vendor catalogs).
The strength of this strategy was supported by preliminary wet-lab validation: guided by its dynamic dictionary, RetroMPA predicted viable, previously unreported substrate combinations for three classic reaction paradigms—a Suzuki–Miyaura coupling, a Bucherer reaction, and a Friedel–Crafts acylation, as illustrated in Fig.~\ref{fig:wet}.
In every case, none of the involved reactants or products existed in the initial dictionary, and the novelty lies exclusively in the predicted reactant pairings.
Each reaction was subsequently executed in the laboratory under standard conditions, and the products were confirmed by $^{1}$H and $^{13}$C NMR spectroscopy; this analysis is detailed in Section S2, Fig.~S2 and Fig.~S3 of the Supporting Information.
These experiments provide preliminary experimental support that RetroMPA can propose synthetically viable substrate pairings beyond the initial dictionary.

From a computational standpoint, our analysis further shows that the retrieval process is both effective and efficient. Using a nearest-neighbor search, the dictionary provides additional chemical constraints to guide accurate predictions, with diminishing returns observed for larger $K$ values, as shown in Table~\ref{tab:knn}. 
The top-K accuracy aligns with standard retrieval-based evaluation in reaction prediction~\cite{xie_retrosynthesis_2023}, where a prediction is considered correct if any of the top-K retrieved candidates matches the ground truth, rather than post-retrieval ranking. From these results, we first observe that incorporating a single retrieval ($K=1$) improves accuracy from 60.34\% to 64.75\%. When $K \geq 5$, accuracy further increases to approximately 65.89\%, with only minor additional gains observed beyond this threshold. We hypothesize this occurs because sufficient constraint information is already captured at $K=5$ to guide the base model's predictions, while molecules farther from the query provide diminishing marginal utility due to their weaker chemical relevance to the target reaction.

In addition to validating its effectiveness, we also conducted a systematic investigation into the computational efficiency of querying the dynamic molecular dictionary. As shown in Table~\ref{tab:efficiency}, we measured the k-NN search time across varying molecular dictionary sizes on an RTX 4090 GPU, using top-K = 50, and a batch size of 32. k-NN search takes 15.01 ms even with a 200K-molecule dictionary. For large-scale deployment, we recommend using the Milvus database for storing and searching molecular vectors. Milvus is a highly optimized vector database system, and benchmarks show search latencies below 0.3 s even for 50 M vectors\cite{milvus_docs}.

As shown in Table~\ref{tab:dict_scale_merged}, we further examined how dictionary scale and composition affect RetroMPA by expanding the original molecular dictionary with randomly sampled decoy molecules from PubChem or ChEBI-20. With PubChem decoys, Top-1 accuracy changed from 64.75\% at 1.0× to 64.49\%, 64.07\%, and 63.61\% at 2.0×, 4.0×, and 8.0×, respectively. With ChEBI-20 decoys, the corresponding accuracies were 64.42\%, 63.93\%, and 63.41\%. These results suggest that performance is affected more by retrieval-space density than by the specific source of added molecules, while RetroMPA still outperforms the EditRetro baseline under all tested expansion settings.

The combination of preliminary wet-lab validation of previously unreported substrate combinations and computationally efficient constrained retrieval establishes RetroMPA as a useful auxiliary component for practical AI-assisted retrosynthesis.  It may help make the model’s outputs more chemically grounded while maintaining computational efficiency.

\begin{table}[t]
\begin{minipage}{0.49\textwidth}  

\centering
\caption{Study on the number of neighbors. }
\label{tab:knn}
\setlength{\tabcolsep}{3pt} 

\begin{tabular}{@{}l *{5}{c} @{}}  
\toprule
Neighbors & 1 & 3 & 5 & 10 & 50 \\
\midrule
Accuracy & 64.75 & 65.79 & 65.89 & 66.01 & 66.17  \\
\bottomrule
\end{tabular}
\end{minipage}
\hfill
\begin{minipage}{0.49\textwidth}
\centering
\caption{Computational efficiency.}
\label{tab:efficiency}

\begin{tabular}{lccc}
\toprule
Molecular dictionary size & 10,000 & 100,000 & 200,000 \\
\midrule
Latency (ms) & 0.51 & 8.38 & 15.01 \\
\bottomrule
\end{tabular}
\end{minipage}
\end{table}

\begin{table}[t]
\centering
\caption{Top-1 accuracy of EditRetro and RetroMPA with different dictionary scale multipliers, where additional molecules are sampled from PubChem (first row) and ChEBI-20 (second row).}
\label{tab:dict_scale_merged}
\begin{tabular}{lccccc}
\toprule
Source & EditRetro (base model) & 1.0x & 2.0x & 4.0x & 8.0x \\
\midrule
PubChem  & 60.34 & 64.75 & 64.49 & 64.07 & 63.61 \\
ChEBI-20 & 60.34 & 64.75 & 64.42 & 63.93 & 63.41 \\
\bottomrule
\end{tabular}
\end{table}

\begin{figure}[t] 
    \centering
    \includegraphics[width=1\textwidth]{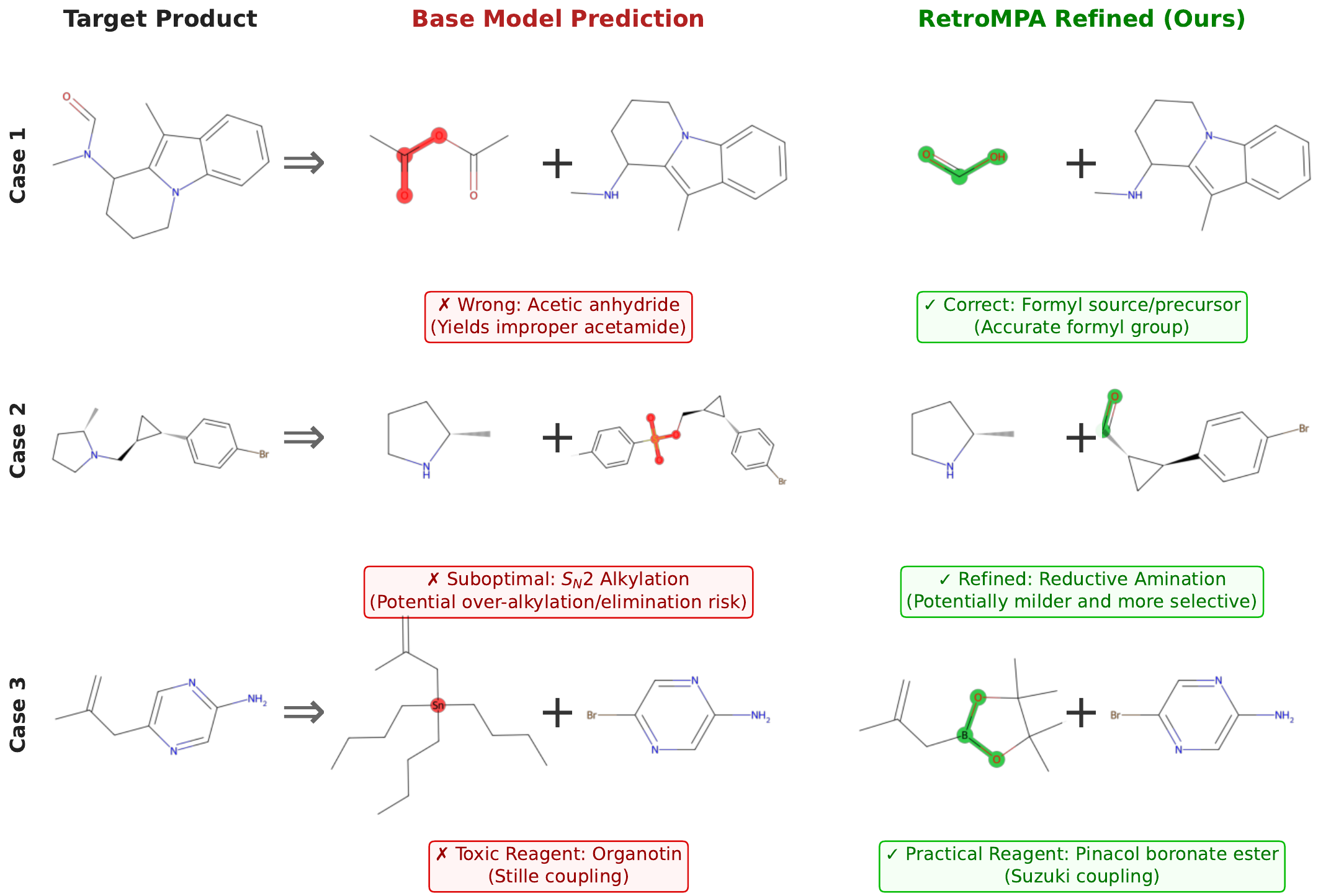}
    \caption{\textbf{Illustrative examples of RetroMPA refinements driven by molecular property awareness.}}
    \label{fig:case_studies}
\end{figure}

\subsection{Case Study: Property-Aware Reasoning for Chemically Interpretable Predictions}
To further elucidate the specific types of molecular properties captured by our framework and to understand their potential advantages, 
we conducted an in-depth chemical analysis of representative cases, as shown in Fig.~\ref{fig:case_studies}. 
These qualitative evaluations aim to demonstrate how the molecular property embeddings may guide context-selective chemical reasoning.

\paragraph{Case 1: Distinguishing Functional Groups and Steric Properties}
\begin{itemize}
    \item \textbf{Product:} \texttt{Cc1c2n(c3ccccc13)CCCC2N(C)C=O}
    \item \textbf{Base Prediction:} \texttt{CC(=O)OC(C)=O} and \texttt{CNC1CCCn2c1c(C)c1ccccc12}
    \item \textbf{RetroMPA Refined:} \texttt{O=CO} and \texttt{CNC1CCCn2c1c(C)c1ccccc12}
\end{itemize}
\textbf{Analysis:} The target product features a specific formamide moiety (\texttt{-N(C)C=O}). The baseline model proposed acetic anhydride (\texttt{CC(=O)OC(C)=O}), an acetylating reagent, which would more plausibly introduce an acetyl group rather than the desired formyl group. From a mechanistic standpoint, this prediction is suboptimal, as it would likely lead to the formation of an acetamide group (\texttt{-N(C)C(=O)CH3}), thereby deviating from the intended target structure. In contrast, RetroMPA refined this prediction to formic acid (\texttt{O=CO}). This adjustment suggests that the incorporation of molecular property embeddings may enhance the model's sensitivity to subtle steric variations and specific functional group definitions (e.g., formyl versus acetyl). Consequently, this property-aware constraint can help mitigate fundamental structural inaccuracies that might be overlooked by purely statistical sequence matching approaches.

\paragraph{Case 2: Recognizing Reaction Selectivity and Strategic Disconnections}
\begin{itemize}
    \item \textbf{Product:} \texttt{C[C@H]1CCCN1C[C@H]1C[C@@H]1c1ccc(Br)cc1}
    \item \textbf{Base Prediction:} \texttt{C[C@H]1CCCN1} and \texttt{Cc1ccc(S(=O)(=O)OC[C@H]2C[C@@H]2c2ccc(Br)cc2)cc1}
    \item \textbf{RetroMPA Refined:} \texttt{C[C@H]1CCCN1} and \texttt{O=C[C@H]1C[C@@H]1c1ccc(Br)cc1}
\end{itemize}
\textbf{Analysis:} Case 2 involves the construction of a sterically hindered C-N bond. The baseline model predicted a direct alkylation pathway employing a bulky tosylate as the electrophile. In practical organic synthesis, direct $S_N2$ alkylation of amines under such sterically encumbered conditions may suffer from over-alkylation or competing elimination. RetroMPA, however, adjusted the prediction toward an aldehyde precursor, effectively redirecting the retrosynthetic strategy to a reductive amination pathway. This refinement indicates that RetroMPA can capture broader chemical contexts regarding reaction selectivity, potentially assisting the model in prioritizing robust and milder synthetic routes for amine formation in complex molecular environments.

\paragraph{Case 3: Differentiating Organometallic Reactivity Profiles}
\begin{itemize}
    \item \textbf{Product:} \texttt{C=C(C)Cc1cnc(N)cn1}
    \item \textbf{Base Prediction:} \texttt{C=C(C)C[Sn](CCCC)(CCCC)CCCC} and \texttt{Nc1cnc(Br)cn1}
    \item \textbf{RetroMPA Refined:} \texttt{C=C(C)CB1OC(C)(C)C(C)(C)O1} and \texttt{Nc1cnc(Br)cn1}
\end{itemize}
\textbf{Analysis:} In Case 3, the baseline model suggested a Stille coupling pathway dependent on an organotin reagent. While theoretically plausible on paper, organotin compounds are often avoided in modern synthetic scaling due to their well-known toxicity and challenging purification profiles. RetroMPA modified this prediction to utilize a pinacol boronate ester, thereby shifting the proposed pathway to a Suzuki–Miyaura coupling. We infer that the multimodal pretraining might equip RetroMPA with an enhanced capacity to differentiate elemental reactivity profiles, effectively reflecting a practical preference for employing more commonly used and generally lower-toxicity organoboron reagents.

\begin{figure}[t] 
\centering 
\includegraphics[width=1\textwidth,trim=0cm 7.5cm 0cm 3.5cm, clip]{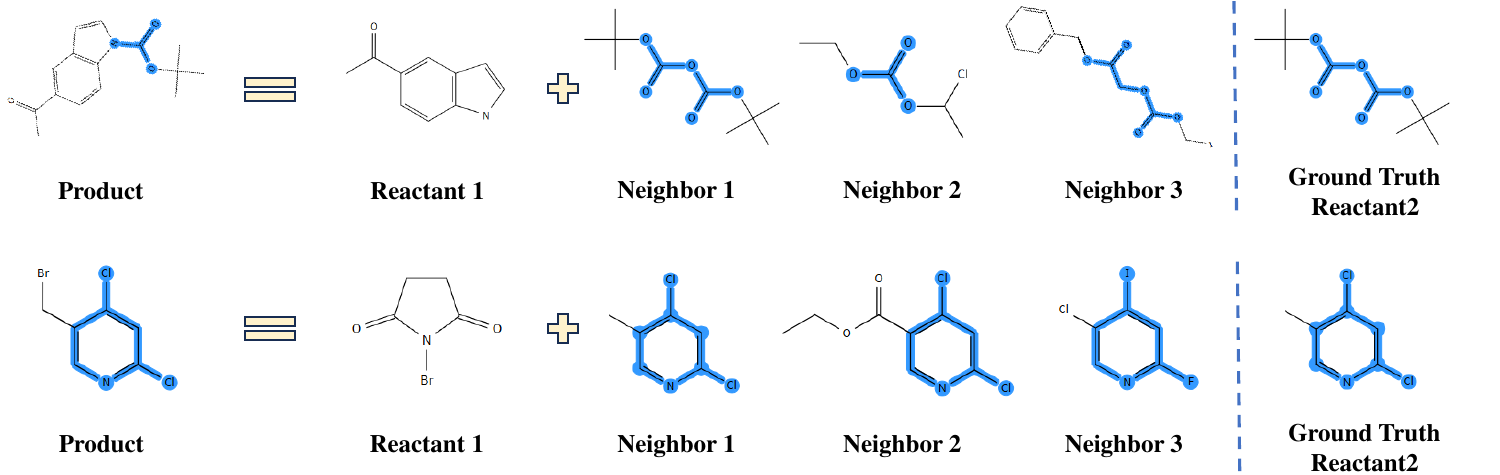} 
\caption{\textbf{Case study of retrieved molecules.}
The columns show the product, the first reactant, the top three nearest-neighbor candidates for the second reactant, and the ground-truth second reactant. Shared substructures are highlighted in blue.
} 
\label{fig:case} 
\end{figure}

We further conducted additional case studies on USPTO-50K by randomly selecting two reactions involving two reactants for intuitive analysis to better understand whether the model can correctly utilize molecular properties in chemical reactions, with results shown in Fig.~\ref{fig:case}. RetroMPA retrieves three candidate nearest-neighbor molecules for the second reactant through inverse projection. The first example (top panel of Fig.~\ref{fig:case}) demonstrates the synthesis of tert-butyl 5-acetylindole-1-carboxylate. RetroMPA retrieves the ground-truth reagent as Neighbor 1, corresponding to Boc protection of the indole N–H functionality through nucleophilic acyl substitution. The blue highlights indicate that other neighbors share the same carbonate ester functional group with ground truth, which could also serve as reactive centers for nucleophilic acyl substitution. The product exhibits a clear N-Boc derivative structure. This demonstrates that RetroMPA  identifies the essential functional group pattern required for the reaction.  Even if the exact reagent is not retrieved, the candidate set remains functionally coherent, allowing a chemist to readily infer the intended transformation.
The second example (bottom panel of Fig.~\ref{fig:case}) illustrates a radical bromination reaction (halogenation reaction) where the product 5-(bromomethyl)-2,4-dichloropyridine retains the structural features of the second reactant – a characteristic of this reaction type. Similarly, the retrieved neighbors share the key heteroaryl with the ground-truth precursor, indicating that RetroMPA retrieves candidates with related reaction-relevant substructures.

These examples suggest that RetroMPA can capture reaction-relevant molecular features that are consistent with selected organic reaction principles. By operating in a molecular property space, the model may generate outputs that are more accurate in the evaluated cases and can provide interpretable chemical cues.     A particularly notable observation is that some retrieved candidates that do not exactly match the ground truth still share reaction-relevant functional groups or substructures and may provide useful chemical clues.  They belong to the same functional group class as the true reactant and could, in principle, participate in analogous reactions under appropriate conditions.  This property-rich output provides valuable intermediate reasoning clues.  Instead of presenting a single black-box prediction, RetroMPA offers a shortlist of chemically plausible options, allowing chemists to infer potential reaction types by analyzing the common property patterns among the candidates.

\begin{figure}[!t] 
\centering

\begin{minipage}[t]{0.49\textwidth} 
\centering
\includegraphics[width=0.96\textwidth,trim=0cm 0.1cm 0cm 0.2cm, clip]{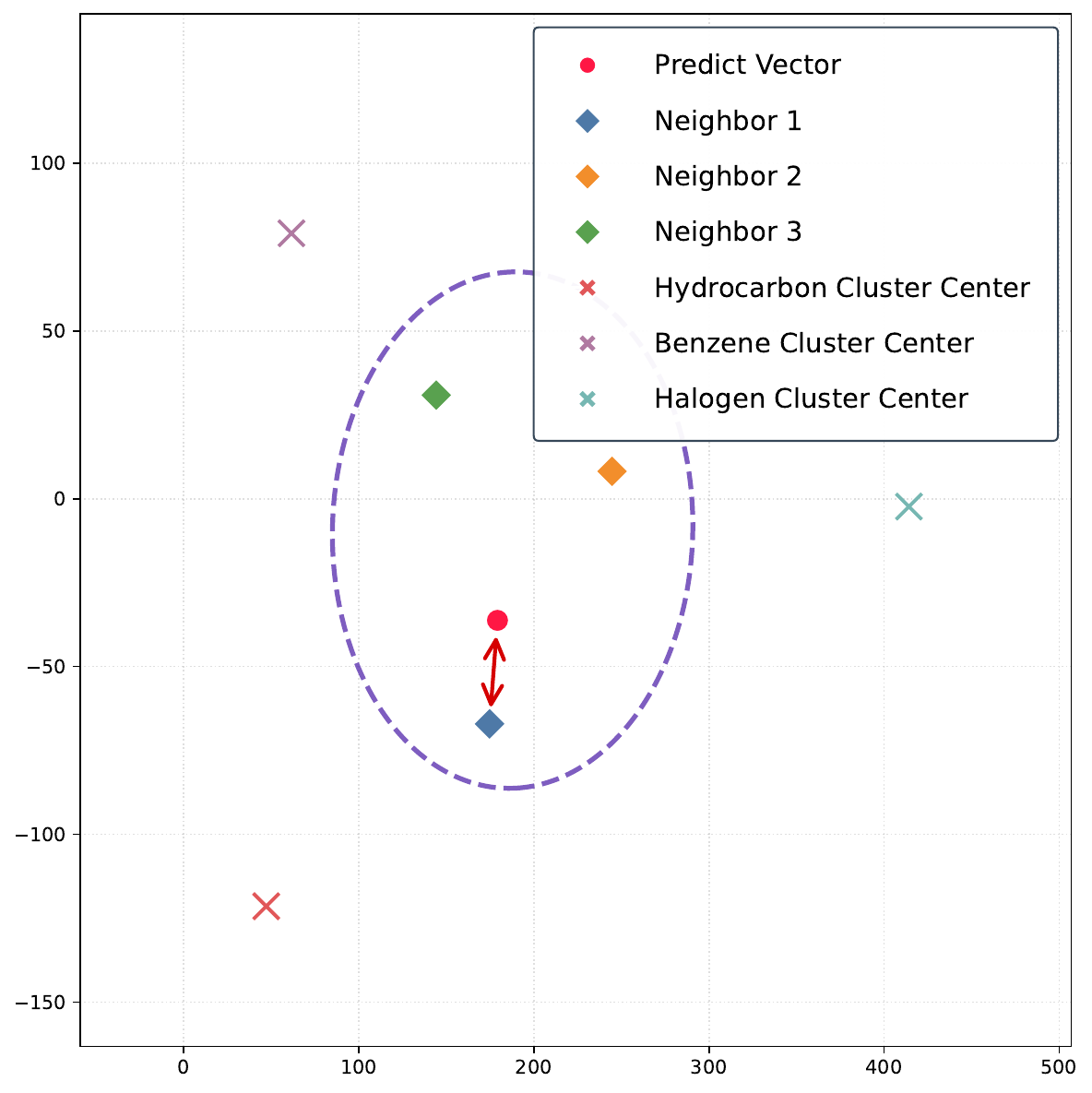} 
\caption{Embedding visualization of predicted vector by RetroMPA.} 
\label{fig:casevisual}
\end{minipage}
\hfill
\begin{minipage}[t]{0.49\textwidth} 
\centering
\includegraphics[width=1\textwidth,trim=0cm 0.1cm 0cm 0.2cm, clip]{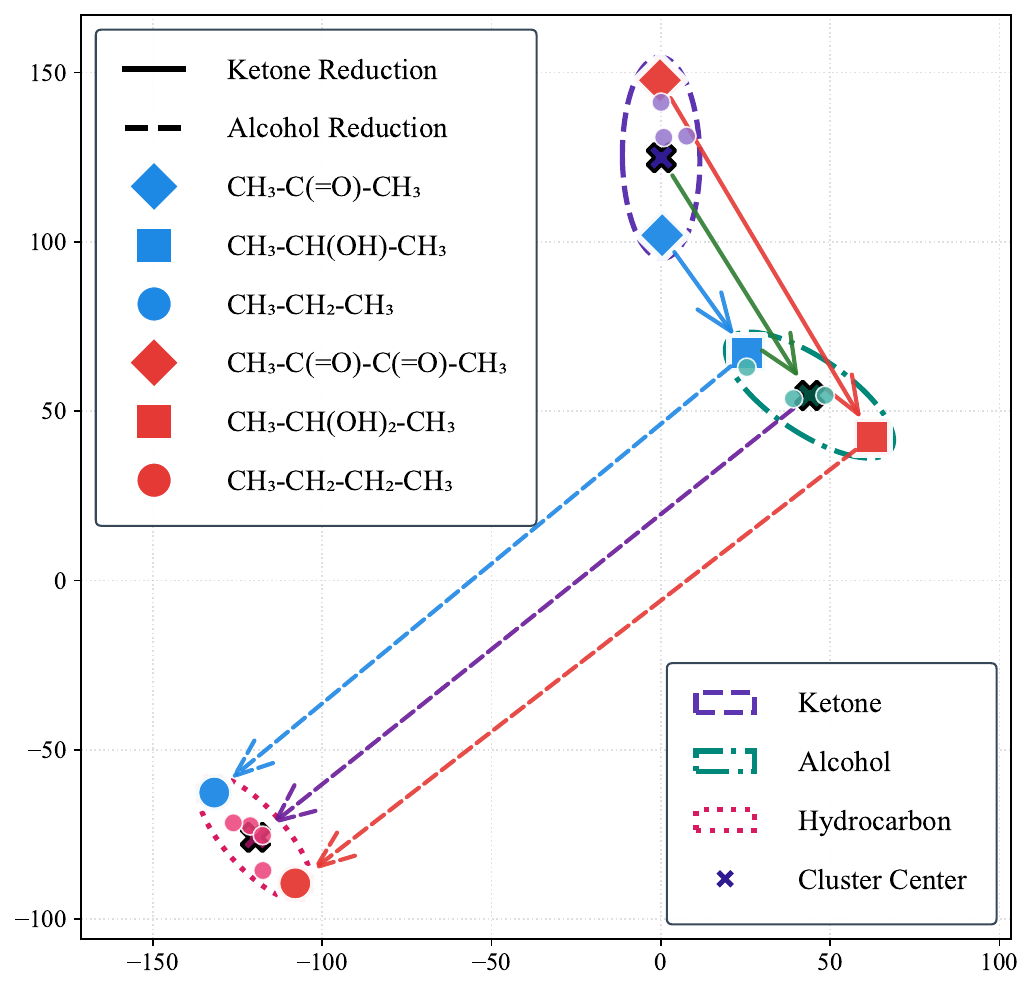} 
\caption{Reaction visualization of ketone reduction and alcohol reduction.} 
\label{fig:visualization1}
\end{minipage}
\end{figure}

\subsection{Embedding visualization}

Further embedding visualization provides deeper insight into how RetroMPA leverages molecular property space. Fig.~\ref{fig:casevisual} shows the embedding visualization for the first case study. The predicted vector for the second reactant in the Boc protection example resides within the \textit{carbonate ester} functional-group cluster and is oriented closer to the \textit{hydrocarbon} region in the embedding space. This geometric alignment indicates that RetroMPA successfully captures the alkane substructure present in the product molecule and correctly attributes it to the second reactant, demonstrating its ability to perform chemically grounded reasoning through property-aware representation learning.

Additionally, we illustrate the importance of molecular properties in chemical reactions through ketone reduction and alcohol reduction examples. 
Their reaction templates are \ce{R_1-CO-R_2 + H_2 -> R_1-CH(OH)-R_2} and 
\ce{R_1-CH(OH)-R_2 ->[Reduction] R_1-CH_2-R_2} respectively.
We first generate molecular property embeddings for propan-2-one (CH\textsubscript{3}COCH\textsubscript{3}) and 2,3-butanedione (CH\textsubscript{3}COCOCH\textsubscript{3}) using the pretrained Mol-Former model, along with corresponding alcohols and hydrocarbons. These vectors are then visualized using t-SNE~\cite{van2008visualizing}, as shown in Fig.~\ref{fig:visualization1}. The results are consistent with: (1) Molecules with similar SMILES but different properties form distinct clusters in Mol-Former's chemical space. This provides embedding-level molecular property priors during retrosynthesis prediction, enabling chemically grounded reasoning. (2) Homologous reactions exhibit analogous directional patterns in the projected low-dimensional property space. This facilitates learning common characteristics and inter-reaction correlations. The model appears to capture functional group quantity effects, as evidenced by the red dashed arrow (representing dual ketone group reduction in 2,3-butanedione) being approximately twice as long as the blue arrow (single ketone reduction). These geometric regularities in chemical space enable more interpretable and chemically plausible prediction compared to traditional black-box approaches.

\section{Conclusion}
In this study, we presented RetroMPA, a knowledge-driven auxiliary framework for retrosynthesis prediction that systematically injects molecular chemical property knowledge to enhance the reliability and performance of existing data-driven models.            Unlike conventional approaches that treat retrosynthesis as a purely statistical sequence generation task, RetroMPA operates at the molecular level, leveraging a chemically grounded embedding space to constrain and refine predictions.          It enables the model to approximate aspects of property-based chemical reasoning used by chemists: not by relying solely on atom-level co-occurrence patterns, but by analyzing how intrinsic molecular properties influence feasible transformations.

Our framework is designed as a plug-and-play module that requires no retraining or architectural modification of existing base models.        This modular design enables seamless post-hoc integration with diverse retrosynthesis architectures without retraining the base models, as demonstrated across eight representative methods spanning template-based, semi-template-based, and template-free paradigms.

Our results demonstrate consistent and substantial improvements in top‑1 accuracy on the USPTO‑50K benchmark, with an average gain of 5.50\% without retraining any base model.            
Even on the more complex and diverse USPTO-Full dataset, the average improvement reaches approximately 2\%, highlighting the framework's ability to leverage chemical knowledge for stronger performance.
Crucially, wet-lab validation provided preliminary evidence that RetroMPA can propose synthetically viable, previously unreported substrate combinations within established reaction classes, including Suzuki–Miyaura coupling, the Bucherer reaction, and Friedel–Crafts acylation.            Case studies further revealed that RetroMPA provides chemically interpretable refinements by retrieving candidates based on shared property patterns, thereby enhancing both prediction accuracy and interpretability.

Despite these advances, several limitations warrant consideration.            RetroMPA is designed as an auxiliary model that requires at least one predicted reactant to initiate its inference process, thus it does not directly apply to single‑reactant reactions.            The inverse projection from continuous embeddings to discrete SMILES via nearest‑neighbor retrieval, while effective, does not guarantee molecular uniqueness and may occasionally retrieve structurally similar but not identical molecules.            Furthermore, the quality of the molecular property embeddings is inherently tied to the breadth and depth of the multimodal pre‑training data.

Looking forward, this work establishes a modular and extensible foundation for knowledge‑augmented retrosynthesis.            The plug‑and‑play nature of RetroMPA makes it readily adaptable to new base models and chemical databases without architectural overhaul.            Several promising directions emerge for future research: (i) iterative application to multi‑step planning, where RetroMPA could refine reaction nodes within retrosynthesis search trees;            (ii) resolving the molecular uniqueness problem through advanced generative decoding or learned inverse mappings;            (iii) integrating reaction condition prediction to provide fully actionable synthetic routes.

Ultimately, by bridging data‑driven machine learning with established chemical principles, RetroMPA represents a step toward more reliable, interpretable, and generalizable AI‑assisted synthesis planning.            We hope this work encourages further integration of domain knowledge as explicit priors in computational chemistry, fostering the development of robust tools that accelerate discovery in organic synthesis and drug development.

\section{Method}
\subsection{Dataset Curation and Preprocessing}

To pre-train the Mol-Former component of RetroMPA for learning molecular chemical properties and thereby guiding retrosynthesis tasks, we leveraged three large-scale multimodal chemical-text datasets:
(i) ChEBI‑20‑MM~\cite{edwards2021text2mol}: contains 33,000 molecule‑description pairs with rich functional‑group and stereochemical annotations.
(ii) Mol‑Instructions~\cite{DBLP:conf/iclr/FangL0LH0FC24}: a biomolecular instruction dataset from which we selected 734,000 chemically relevant entries.
(iii) PubChemSTM~\cite{liu2023multi}: comprises over 280,000 molecule–text pairs describing broader chemical properties and roles (see Fig. S1).

All benchmark evaluations in this work are conducted on the USPTO‑50K dataset~\cite{schneider2016s} and USPTO-Full dataset\cite{dai2020retrosynthesispredictionconditionalgraph}.
USPTO‑50K is a widely adopted standard comprising 50,016 atom-mapped single-step reactions extracted from United States patent literature. 
The USPTO-Full dataset, comprising approximately one million reactions, is substantially larger and more diverse, allowing us to validate our model's performance beyond standard benchmarks. Experiments on both datasets enable a comprehensive assessment of our method’s performance, generalization ability, and scalability, providing valuable insights for practical retrosynthesis applications.
Each reaction is originally represented in SMILES format as reactants > reagents > products.                 
To focus on the core structural transformation and align with common practice in recent retrosynthesis literature, we remove reagent information entirely.                 
Reagents—such as catalysts, solvents, or bases—are generally not intended to contribute atoms to the final product and are often inconsistently annotated across patents or preprocessing pipelines, which introduces ambiguity in role assignment and potential noise during training~\cite{doi:10.1021/acs.jcim.5c01508}.                 
By discarding reagents, our input format simplifies to reactants → product, where all reactants are concatenated using the canonical dot ('.') delimiter.
Considering the noise introduced during dataset construction and the lack of reaction condition descriptions, we retained only reactions with two reactants.

To ensure data integrity and prevent unintended information leakage, we canonicalize all SMILES strings using RDKit~\cite{landrum2013rdkit} and then remove atom mappings and reaction class labels  prior to model training or evaluation to avoid providing the model with explicit atom‑correspondence information unavailable in real‑world prediction scenarios.        
Notably, the molecular dictionary itself is a purely  passive database utilized exclusively during the inference phase (Stage 3). It is not used for training, gradient computation, representation learning, model selection, or hyperparameter tuning, and therefore does not alter the learned model or the generated query representation. Importantly, the dictionary is used only after RetroMPA has generated the query embedding.  For benchmark evaluation, the inverse-projection step is evaluated under a closed-candidate retrieval protocol. The molecular dictionary is constructed as a fixed candidate gallery from molecules in the evaluation candidate pool, including the ground-truth precursor molecules.  This setting is necessary for retrieval-style evaluation because the correct molecule must be present in the candidate gallery to make Hits@k, Recall@k, MRR, or exact-match retrieval accuracy meaningful. This is standard mathematical practice\cite{edwards2021text2mol,bushuiev2024massspecgym,liu2023molca} in retrieval-augmented generation architectures and does not constitute leakage.

All SMILES strings used in both pretraining and downstream evaluation are processed using standard cheminformatics tools.   No SMILES augmentation or standardization beyond canonicalization is applied, and no atom-mapping-based filtering is used, thereby preserving more of the native chemical diversity of the source data.     This curation strategy is intended to approximate realistic prediction conditions by avoiding explicit privileged signals, while allowing the model to benefit from large-scale chemical knowledge during pretraining.

\subsection{Methodological Foundation}

We propose RetroMPA, consisting of a Mol-Former and a Molecular Decoder, which leverages the fundamental principle that chemical reactions inherently depend on molecular properties.  The core idea of RetroMPA is to constrain the retrosynthesis process using the Molecular Decoder through chemically-informed embeddings from Mol-Former. Mol-Former maps molecules into vectors with explicit molecular-property semantics; the Molecular Decoder then predicts reactant embeddings based on these vectors, and the corresponding molecules are obtained through \textit{Inverse Projection}, preserving molecular knowledge priors throughout the retrosynthesis workflow. During the inference stage, we take the product and one reactant predicted by the base model as the starting input, and predict the remaining reactant via RetroMPA, thereby obtaining a complete reaction pathway constrained by chemical knowledge. To formalize this auxiliary role, we define the auxiliary retrosynthesis task.

\textbf{Auxiliary Retrosynthesis.}
We formally define the auxiliary retrosynthesis task as follows. Whereas standard retrosynthesis infers all potential reactants from a given product, auxiliary retrosynthesis predicts the remaining reactants using both the product and one reactant predicted by the base model, thereby refining the base model’s output.
Given product $P$ and the first predicted reactant $R_1$ from a base model, the auxiliary task aims to predict remaining reactants $\{R_2,...,R_n\}$ through eq~\ref{eq1}:
\begin{equation}
\label{eq1}
\{R_2,...,R_n\} = \mathop{\arg\max}\limits_{\mathcal{R}} \Phi(P,R_1,\mathcal{R}|\Theta),
\end{equation}
where $\Phi(\cdot)$ denotes the model parameterized by $\Theta$, $\mathcal{R}$ denotes the candidate solution space for retrosynthesis predictions.

\subsection{Mol-Former: Cross-Modal Molecular Chemical Knowledge Learning}

\textbf{Injecting Chemical Knowledge via Multimodal Pretraining.} 
RetroMPA leverages \textbf{multimodal pretraining} for enhanced robustness. Unlike SMILES-only methods, our framework is pretrained on large-scale chemical-text datasets, equipping it with broader chemical intuition to recognize and appropriately respond to phenomena like functional-group identity and reagent-relevant molecular features.

We introduce \textbf{Mol-Former}, a multimodal encoder that aligns molecular and textual representations through a shared embedding space. It comprises a Molecular Transformer, a Text Transformer, and a Molecular Encoder. The architecture follows the Q-Former design~\citep{li2023blip}, where a learnable Molecular Transformer attends to facilitate interaction between molecular representations and textual chemical knowledge, ultimately transforming the query tokens into molecular property embeddings $\mathbf{Q}_m$ with explicit chemical semantics.
Formally, given a SMILES string and its associated text description, we first encode them into latent sequences using pretrained molecular encoder MolR \citep{wang2021chemical} and a Text Transformer. A learnable query then attends to both modalities via cross-attention, yielding a joint embedding space. To align molecular structures with their semantic descriptions, we train Mol-Former using all three original pre-training objectives (i.e., image-text contrastive, image-text matching, and language modeling loss), with the original image modality simply replaced by molecules, yielding the corresponding molecule–text contrastive (MTC), molecule–text matching (MTM), and language modeling (LM) loss terms. 
Specifically, we optimize eq~\ref{eq2}:
\begin{equation}
\label{eq2}
\mathcal{L}_{\mathrm{Mol\text{-}Former}}
=
\mathcal{L}_{\mathrm{MTC}}
+
\mathcal{L}_{\mathrm{MTM}}
+
\mathcal{L}_{\mathrm{LM}}.
\end{equation}
where $\mathcal{L}_{\text{MTC}}$ is a molecule–text contrastive loss encouraging similar embeddings for matched pairs, $\mathcal{L}_{\text{MTM}}$ promotes accurate identification of positive pairs, and $\mathcal{L}_{\text{LM}}$ enables generation of chemically meaningful descriptions conditioned on the molecule.

This cross-modal alignment objective encourages the model to ground textual semantics, such as ``carboxylic acid'', ``R-enantiomer'', or ``Ketone'', into the structure-derived molecular representation, equipping it with broader chemical intuition to recognize phenomena like functional-group identity and reagent-relevant molecular features. Ultimately Mol-Former transforms the query tokens into molecular property embeddings $\mathbf{Q}_m$ with explicit chemical semantics, as shown in eq~\ref{eq3}:
\begin{equation}
\label{eq3}
\mathbf{Q}_m = \mathrm{Mol\text{-}Former}(\mathbf{m}),
\end{equation}
where $\mathbf{m}$ denotes molecular embeddings generated by molecular encoder. This integration of chemical knowledge into $\mathbf{Q}_m$ enables molecular chemical properties to be used as \textit{prior information} for retrosynthesis prediction.

\subsection{Molecular Decoder: Molecular-Level Retrosynthesis Prediction}\label{sec:stage2}
\textbf{Molecular-Level Prediction.} To better capture the compositional nature of chemical reactions, we formulate retrosynthesis as a  sequence generation task at the molecular level, where each step generates a complete reactant molecule rather than individual atoms, promoting higher-level chemical reasoning. We define the solution retrosynthesis prediction space $\mathcal{R}$ as a sequence of molecular mappings $(m_1, \dots, m_K)$, each representing a reactant in the context of the product. Given the product embedding $\mathbf{Q}_{pro}$ from Mol-Former and previous predictions, the Molecular Decoder models the conditional likelihood, as given in eq~\ref{eq4}:
\begin{equation}
\label{eq4}
    P(\mathcal{R}) = \prod_{k=1}^K P(m_k \mid m_{i<k}, \mathbf{Q}_{pro}),
\end{equation}
where \( m_k \) denotes the \( k \)-th reactant. During training, we ground predictions on ground-truth reactants from standard datasets to avoid potential bias or performance limitations that might arise from relying on any base model's predictions, which enables universal compatibility across diverse predictors. We employ teacher forcing~\citep{feng2021guiding}, conditioning each step on true preceding reactants.

\textbf{Learning Discriminative Reactant Representations via Dual-Prior Contrastive Learning.} 




To ensure the Molecular Decoder generates both accurate and chemically meaningful reactants, we design a contrastive learning framework. Let $\hat{\mathbf{r}}_i^{(k)}$ denote the predicted embedding of the $k$-th reactant for sample $i$, and let $\mathbf{m}_i^{(k)}$ denote the corresponding ground-truth molecular embedding encoded by the molecular encoder. The Decoder is optimized by combining Molecular Contrastive Loss (MCL) and Molecular Matching Loss (MML).

\begin{itemize}
\item \textbf{Molecular Matching Loss (MML):}
First, a \textbf{M}olecular \textbf{M}atching \textbf{L}oss (MML) attempts to minimize the L2 distance between predicted reactant embeddings $\mathbf{r}$ and their targets $\mathbf{m}$. 
The positive matching term given in eq~\ref{eq5} minimizes the L2 distance between each predicted reactant embedding and its corresponding ground-truth embedding:
    \begin{equation}
    \label{eq5}
    \mathcal{L}_{\mathrm{pos}}
    =
    \lambda_1
    \sum_{i=1}^{B}
    \left\|
    \hat{\mathbf{r}}_i^{(1)}
    -
    \mathbf{m}_i^{(1)}
    \right\|_2
    +
    \lambda_2
    \sum_{i=1}^{B}
    \left\|
    \hat{\mathbf{r}}_i^{(2)}
    -
    \mathbf{m}_i^{(2)}
    \right\|_2 .
    \end{equation}

To prevent representation collapse and encourage separation from difficult negatives, we sample one hard negative for each sample and each reactant type. Specifically, the positive diagonal entry is masked out, and the negative index is sampled from a multinomial distribution derived from the softmax-normalized prediction-to-reactant similarities. Let $\mathbf{m}_{n_i^{(k)}}^{(k)}$ denote the sampled negative embedding for the $k$-th reactant of sample $i$. The negative matching term is given in eq~\ref{eq6}:
    \begin{equation}
    \label{eq6}
    \mathcal{L}_{\mathrm{neg}}
    =
    \lambda_1
    \sum_{i=1}^{B}
    \left\|
    \hat{\mathbf{r}}_i^{(1)}
    -
    \mathbf{m}_{n_i^{(1)}}^{(1)}
    \right\|_2
    +
    \lambda_2
    \sum_{i=1}^{B}
    \left\|
    \hat{\mathbf{r}}_i^{(2)}
    -
    \mathbf{m}_{n_i^{(2)}}^{(2)}
    \right\|_2 .
    \end{equation}

    The Molecular Matching Loss is then defined as shown in eq~\ref{eq7}:
    \begin{equation}
    \label{eq7}
    \mathcal{L}_{\mathrm{MML}}
    =
    \frac{
    \mathcal{L}_{\mathrm{pos}}
    -
    \mathcal{L}_{\mathrm{neg}}
    }{2}.
    \end{equation}
\end{itemize}

\begin{itemize}
    \item \textbf{Molecular Contrastive Loss (MCL):}

Second, standard contrastive learning often inadvertently penalizes false negatives—molecules that are structurally or functionally viable alternatives within the batch. To mitigate this limitation, we draw inspiration from recent advancements in contrastive learning, specifically iMolCLR\cite{wang2022improving} and ProGCL\cite{xia2022progcl}, to formulate a re-weighted \textbf{M}olecular \textbf{C}ontrastive \textbf{L}oss (MCL) enhanced by a dual-prior mechanism. We adjust the penalty of negative samples by injecting two weighting factors. 
For each reactant type, we compute bidirectional similarities between the predicted reactant embeddings and the ground-truth reactant embeddings, as given in eq~\ref{eq8}:
    \begin{equation}
    \label{eq8}
    s_{ij}^{r2p,(k)}
    =
    \frac{
    \mathbf{m}_i^{(k)\top}\hat{\mathbf{r}}_j^{(k)}
    }{\tau},
    \quad
    s_{ij}^{p2r,(k)}
    =
    \frac{
    \hat{\mathbf{r}}_i^{(k)\top}\mathbf{m}_j^{(k)}
    }{\tau},
    \end{equation}
    where $\tau$ is a learnable temperature parameter initialized to 0.07.
To reduce the influence of potential false negatives in the batch, we apply a dual-prior weighting strategy to negative logits. First, following the structural prior used in iMolCLR\cite{wang2022improving}, we use the Tanimoto similarity between molecular fingerprints to down-weight structurally similar negatives, as shown in eq~\ref{eq9}:
    \begin{equation}
    \label{eq9}
    w_{ij}^{\mathrm{struc},(k)}
    =
    1-\mathrm{Tanimoto}
    \left(
    \mathbf{fp}_i^{(k)}, \mathbf{fp}_j^{(k)}
    \right).
    \end{equation}
    The fingerprints are Morgan fingerprints generated from SMILES strings with radius 2 and 2048 bits. Invalid SMILES are represented by zero vectors. This down-weights negatives that have high 2D structural similarity to the query, as they may possess similar chemical properties, mitigating the effect of false negatives in the batch.

Then, following the probabilistic prior used in ProGCL\cite{xia2022progcl}, we do not explicitly select true negatives with hard labels. Instead, for each in-batch candidate negative, we estimate its posterior probability of being a true negative from the batch-level cosine similarity distribution,as given in eq~\ref{eq10}:
    \begin{equation}
    \label{eq10}
    w_{ij}^{\mathrm{prob},(k)}
    =
    P(\mathrm{True\ Negative}\mid \mathrm{sim}_{ij}).
    \end{equation}
    In implementation, this probability is obtained by fitting a two-component Beta mixture model to the normalized cosine similarities in the current batch.

    Eq~\ref{eq11} is the final weight for each negative sample:
    \begin{equation}
    \label{eq11}
    w_{ij}^{(k)}
    =
    \max
    \left(
    w_{ij}^{\mathrm{struc},(k)}
    \cdot
    w_{ij}^{\mathrm{prob},(k)},
    10^{-6}
    \right).
    \end{equation}
    The weight is injected into the logits only for negative samples, as given in eq~\ref{eq12}:
    \begin{equation}
    \label{eq12}
    \tilde{s}_{ij}^{(k)}
    =
    \begin{cases}
    s_{ij}^{(k)}, & j=y_i,\\
    s_{ij}^{(k)}+\log w_{ij}^{(k)}, & j\neq y_i,
    \end{cases}
    \end{equation}
    where $y_i$ is the positive label corresponding to the matched sample in the batch.

It provides a probabilistic assessment of whether each candidate negative is a true negative. It dynamically fits a two-component Beta mixture model to the distribution of cosine similarities within the current batch. The posterior probability of a sample belonging to the low-similarity (true negative) component is used as its weight. This adaptively down-weights potential false negatives that have unexpectedly high representation similarity. This weighting strategy encourages the model to distinguish distinct molecules without overly penalizing chemically reasonable alternatives.
\end{itemize}

The final Molecular Contrastive Loss given in eq~\ref{eq13} is the weighted average of four cross-entropy terms, covering both directions and both reactants:
    \begin{equation}
    \label{eq13}
    \mathcal{L}_{\mathrm{MCL}}
    =
    \frac{1}{4}
    \left(
    \lambda_1
    \mathrm{CE}_{\epsilon}
    \left(
    \tilde{S}_{r2p}^{(1)}, \mathbf{y}
    \right)
    +
    \lambda_2
    \mathrm{CE}_{\epsilon}
    \left(
    \tilde{S}_{r2p}^{(2)}, \mathbf{y}
    \right)
    +
    \lambda_1
    \mathrm{CE}_{\epsilon}
    \left(
    \tilde{S}_{p2r}^{(1)}, \mathbf{y}
    \right)
    +
    \lambda_2
    \mathrm{CE}_{\epsilon}
    \left(
    \tilde{S}_{p2r}^{(2)}, \mathbf{y}
    \right)
    \right),
    \end{equation}
    where $\lambda_1$ and $\lambda_2$ are the weights for reactant 1 and reactant 2, respectively, and $\mathrm{CE}_{\epsilon}$ denotes cross-entropy with label smoothing $\epsilon=0.1$. All embeddings are L2-normalized before similarity computation, so the dot product corresponds to cosine similarity.

Training proceeds in two phases: (1) Mol-Former is frozen while the Molecular Decoder focuses purely on learning correlations between product and reactant properties without interference from encoder updates. (2) End-to-end fine-tuning refines both Mol-Former and Molecular Decoder. The overall objective is shown in eq~\ref{eq14}:

\begin{equation}
\label{eq14}
\mathcal{L}_{\mathrm{Decoder}}
=
\mathcal{L}_{\mathrm{MCL}}
+
\mathcal{L}_{\mathrm{MML}}.
\end{equation}
guiding the model to learn the chemical property relationships between the product and its corresponding reactants. Furthermore, our property-based filtering approach attempts to provide a fundamentally less biased evaluation toward rare but valid reaction pathways. Unlike purely data-driven methods, which often struggle to learn underrepresented pathways due to limited training examples, our framework evaluates reactions based on reactants' intrinsic chemical properties rather than statistical prevalence. We hope this mechanism may reduce the tendency to overlook rare but chemically plausible pathways by incorporating property-based constraints, regardless of their frequency, thereby helping RetroMPA to avoid ignoring valid but rare reaction pathways.

\subsection{Inverse Projection and Molecular Output}
The reconstruction of SMILES strings from vector representations \( \mathbf{r} \) poses non-trivial challenges. First, the solution retrosynthesis prediction space \( \mathcal{R} \) and the ground-truth molecular embedding space are inherently non-equivalent, as they originate from distinct model outputs. Second, \( \mathbf{r} \) predicted by the Molecular Decoder cannot maintain precise one-to-one correspondence with actual molecules. To resolve this issue and enable deterministic molecular output in SMILES format, we introduce a Molecular Dictionary. 

\textbf{Molecular Dictionary.}
The molecular dictionary serves as a fundamental key-value store, wherein the values correspond to molecular embeddings, and the keys are the associated molecular SMILES strings. For benchmarking baseline performance on standard datasets, the initial dictionary is populated with molecules drawn exclusively from the dataset. However, real-world chemical synthesis is not confined to the boundaries of any fixed dataset. Consequently, the molecular dictionary is designed to be dynamic rather than static. This dynamic dictionary automatically incorporates candidate molecules predicted by the base model into its repository.   This mechanism enables RetroMPA to consider candidate molecules absent from the original dictionary once they are added from base-model outputs.   In practical deployment scenarios, the molecular dictionary can also be manually augmented in bulk with catalogs of molecules from real-world chemical vendors (e.g., Enamine). The molecular dictionary itself is purely a passive database utilized exclusively during the Stage 3 inference phase, it does not participate in backpropagation or training.

Formally, given a predicted embedding $\mathbf{r}$, we retrieve the most chemically similar molecule $M_p$ from a dictionary $\mathcal{D}$ of known compounds by maximizing a composite similarity score. To capture both the directional alignment and the spatial proximity of the embeddings, our scoring function combines cosine similarity and a normalized Euclidean distance.

Specifically, for a dictionary molecule $M \in \mathcal{D}$ with embedding $\mathbf{m}_M$, the cosine similarity is defined as:
\begin{equation}
S_{cos}(\mathbf{r}, \mathbf{m}_M) = \frac{\mathbf{r} \cdot \mathbf{m}_M}{|\mathbf{r}|_2 |\mathbf{m}_M|_2}
\end{equation}
where $\mathbf{m}_M$ is the embedding of molecule $M$, computed in the same embedding space as the Molecular Decoder output.
For the Euclidean distance $d(\mathbf{r}, \mathbf{m}_M) = \|\mathbf{r} - \mathbf{m}_M\|_2$, we apply a min-max normalization across all molecules in the dictionary to map the distances into a $[0, 1]$ range. The normalized distance is then inverted to form a distance-based similarity metric $S_{dist}$:
\begin{equation}
S_{dist}(\mathbf{r}, \mathbf{m}_M) = 1 - \frac{d(\mathbf{r}, \mathbf{m}_M) - d{min}}{d{max} - d_{min} + \epsilon}
\end{equation}
where $d_{min} = \min_{M' \in \mathcal{D}} d(\mathbf{r}, \mathbf{m}_{M'})$ and $d_{max} = \max_{M' \in \mathcal{D}} d(\mathbf{r}, \mathbf{m}_{M'})$ denote the minimum and maximum Euclidean distances for the current query $\mathbf{r}$ over the entire dictionary $\mathcal{D}$, and $\epsilon = 10^{-8}$ is a small constant added for numerical stability.

Finally, the target molecule $M_p$ is retrieved by maximizing the sum of these two similarities:
\begin{equation}
M_p = \argmax_{M \in \mathcal{D}} \Big( S_{cos}(\mathbf{r}, \mathbf{m}_M) + S_{dist}(\mathbf{r}, \mathbf{m}_M) \Big)
\end{equation}

For Top-$K$ retrieval, we select the $K$ molecules that yield the highest combined scores.
This nearest-neighbor search helps return valid molecules from the dictionary and favors candidates that are structurally compatible with the predicted chemical properties. The specific output algorithm is presented in Algorithm S1.

\subsection{Training}
As aforementioned, we adopt the original Q-Former architecture and MolR ~\cite{wang2021chemical} framework to implement Mol-Former. MolR is initialized with weights from its official GitHub repository. We employ a minimalist design, a standard transformer decoder ~\cite{vaswani2017attention} with 6 layers and 8 attention heads as our Molecular Decoder, but remove the final softmax layer to enable direct vector output. We set 192 as the maximum text token length and use AdamW optimizer ~\cite{DBLP:conf/iclr/LoshchilovH19} with the peak learning rate $1\times10^{-3}$. Pretraining was conducted for 351K steps on four NVIDIA RTX 4090 GPUs. The Molecular Decoder was then trained without reaction class labels for 200 epochs using the AdamW optimizer with a learning rate of $1\times10^{-4}$.

\section{Data and Software Availability}
The dataset used in this study and the code files used to perform the experiments can be found at \href{https://github.com/MengzhouLu/RetroMPA}{https://github.com/MengzhouLu/RetroMPA}.

\section*{Author Contributions}
†M.L. and F.X. contributed equally to this work. Y.W. conceptualized and designed the study, supervised the research, acquired funding, and critically reviewed and edited the manuscript. M.L. and F.X. established the methodological framework, designed the experiments, performed formal data analysis, and wrote the original draft. M.L. additionally developed and validated the computational code. Z.Y. prepared the experimental figures and visualizations. Z.Y. and H.H. provided writing guidance and reviewed and edited the manuscript. Y.L. and Y.Y. provided mathematical analysis support. W.L. provided chemical expertise and conducted the wet-lab experimental validation. All authors have given approval to the final version of the manuscript.

\section*{Notes}
The authors declare no competing financial interests.

\section*{Acknowledgements}

This work was partially supported by the National Science and Technology Major Project (2023ZD0120802), Wuhan Key Research and Development Program (Grant No. 2025051202030408) and the Fundamental Research Funds for the Central Universities (2042026kf0055).

\section*{Supporting Information}


\begin{itemize}
  \item SI.pdf: This Supporting Information includes dataset descriptions, experimental NMR spectral validation, and the RetroMPA inference algorithm (PDF).
\end{itemize}


\bibliography{JCIM}

\newpage
\begin{center}
    \Large\textbf{TOC Graphic} \\[1cm] 
    
    \includegraphics[width=1\textwidth]{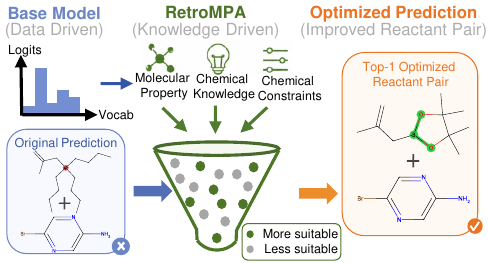} 
\end{center}
\end{document}


\maketitle

\section*{Contents}

Fig.~\ref{fig:datasets} illustrates the data format of the chemical datasets used for pretraining. Fig.~\ref{fig:Hpu} and Fig.~\ref{fig:Cpu} present the experimental ¹H and ¹³C NMR spectra of the wet-lab product, respectively. Algorithm.~\ref{alg:infer} details the full inference procedure of the framework.

\section*{S1.Analysis of the chemical datasets used for pretraining}
\begin{figure}[t] 
\centering 
\includegraphics[width=1\textwidth,trim=1cm 2.5cm 1cm 2.5cm, clip]{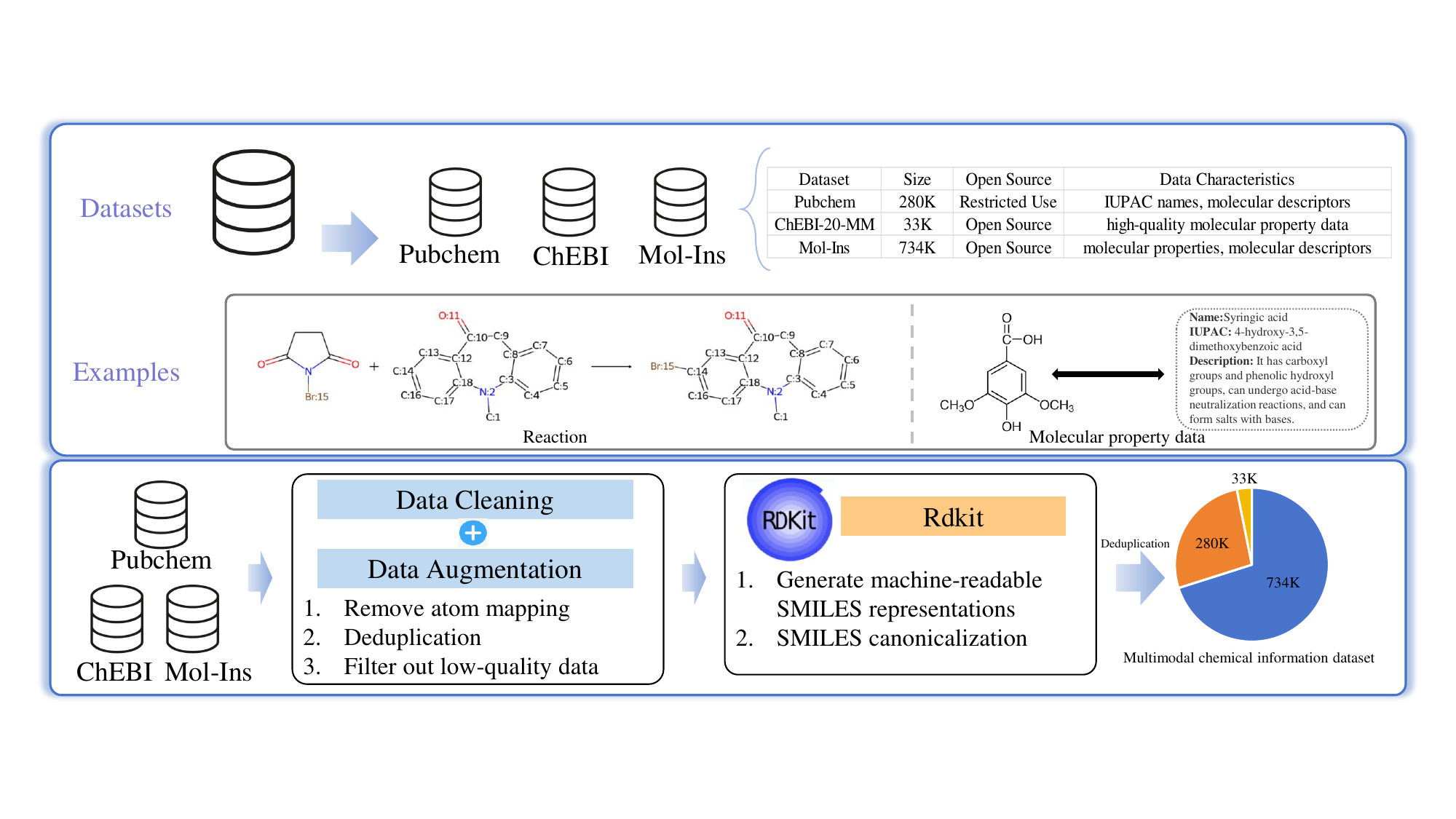} 
\caption{Introduction of datasets.}
\label{fig:datasets} 
\end{figure}

To pretrain the Mol-Former component of RetroMPA for learning molecular chemical properties and thereby guiding retrosynthesis tasks, we leveraged three large-scale chemical-text datasets: ChEBI‑20‑MM, Mol‑Instructions, and PubChemSTM, collectively endowing the model with rich molecular knowledge, see Fig.~\ref{fig:datasets}.
Together, these sources provide rich semantic grounding: a portion of the text explicitly describes critical chemical attributes (e.g., functional groups, chirality), while the majority cover broader molecular features such as electronic properties or biochemical roles. This multimodal pretraining enables Mol-Former to align molecules with natural language descriptions, producing embeddings enriched with interpretable chemical semantics.

\section{S2.Experimental chemical analysis}

\begin{figure}[!t]
    \centering
    \includegraphics[width=0.32\linewidth]{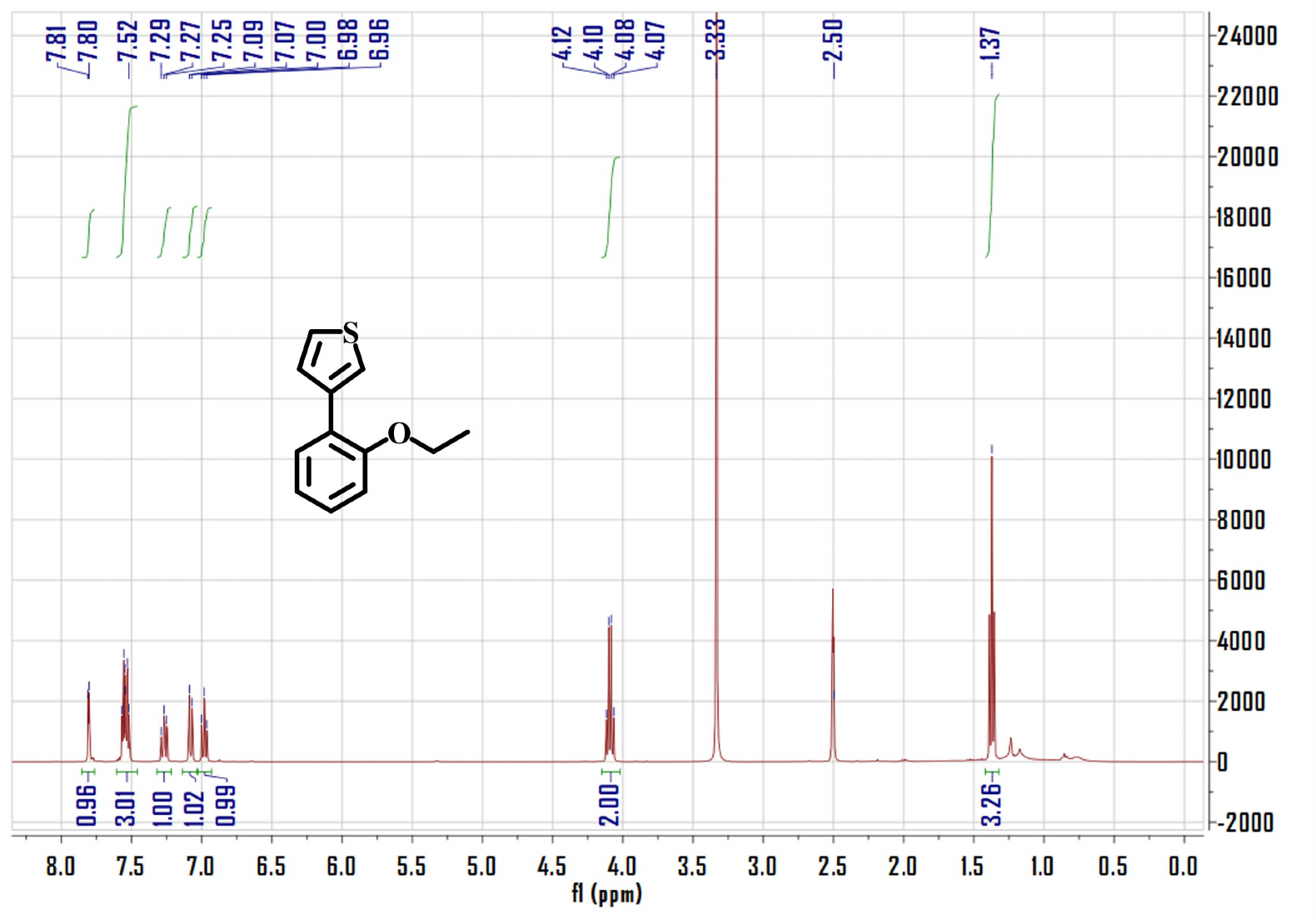}\hfill
    \includegraphics[width=0.32\linewidth]{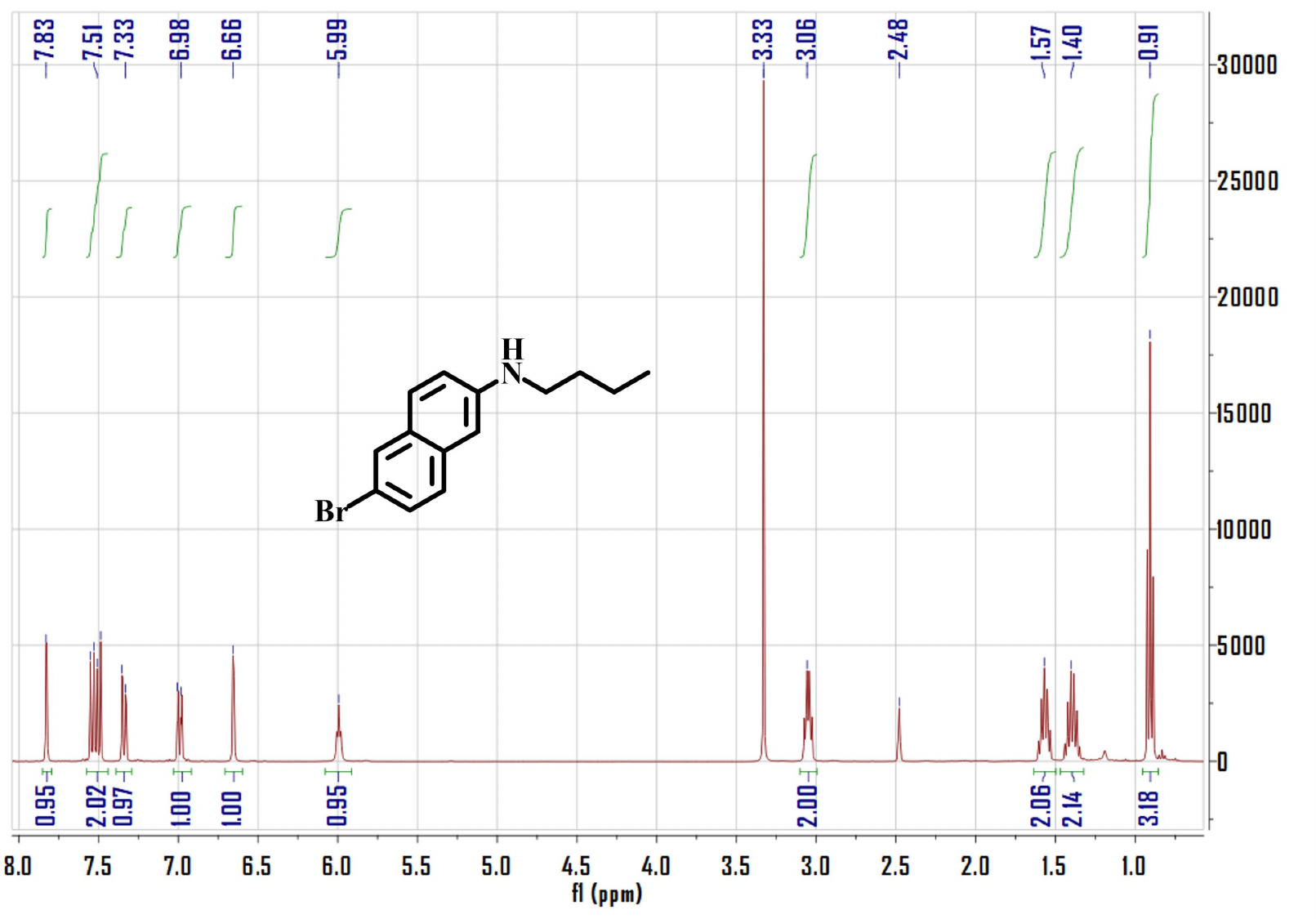}\hfill
    \includegraphics[width=0.32\linewidth]{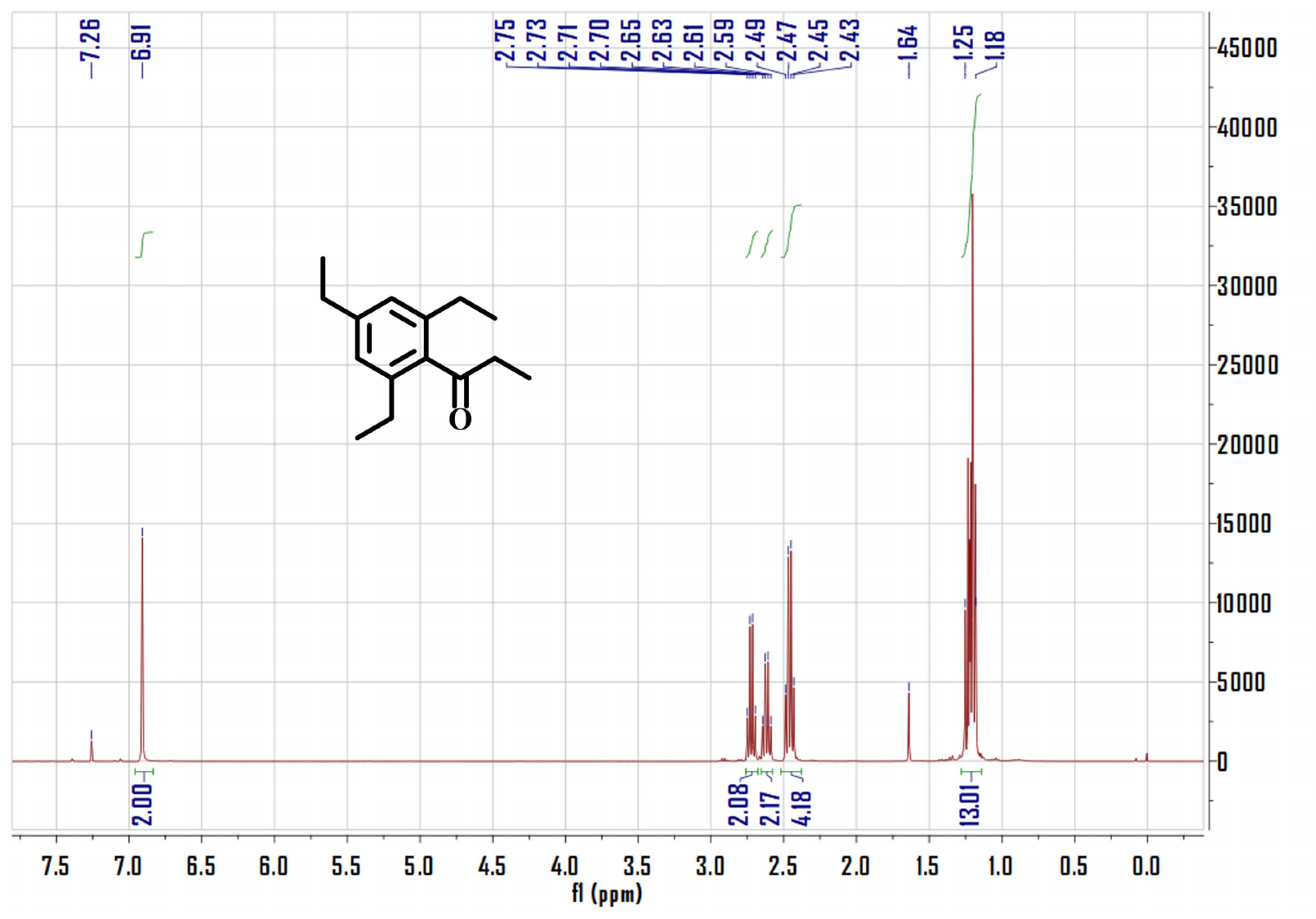}
    \caption{¹H NMR spectra of all chemical products. \label{fig:Hpu}}
    
    \vspace{1em} 
    \includegraphics[width=0.32\linewidth]{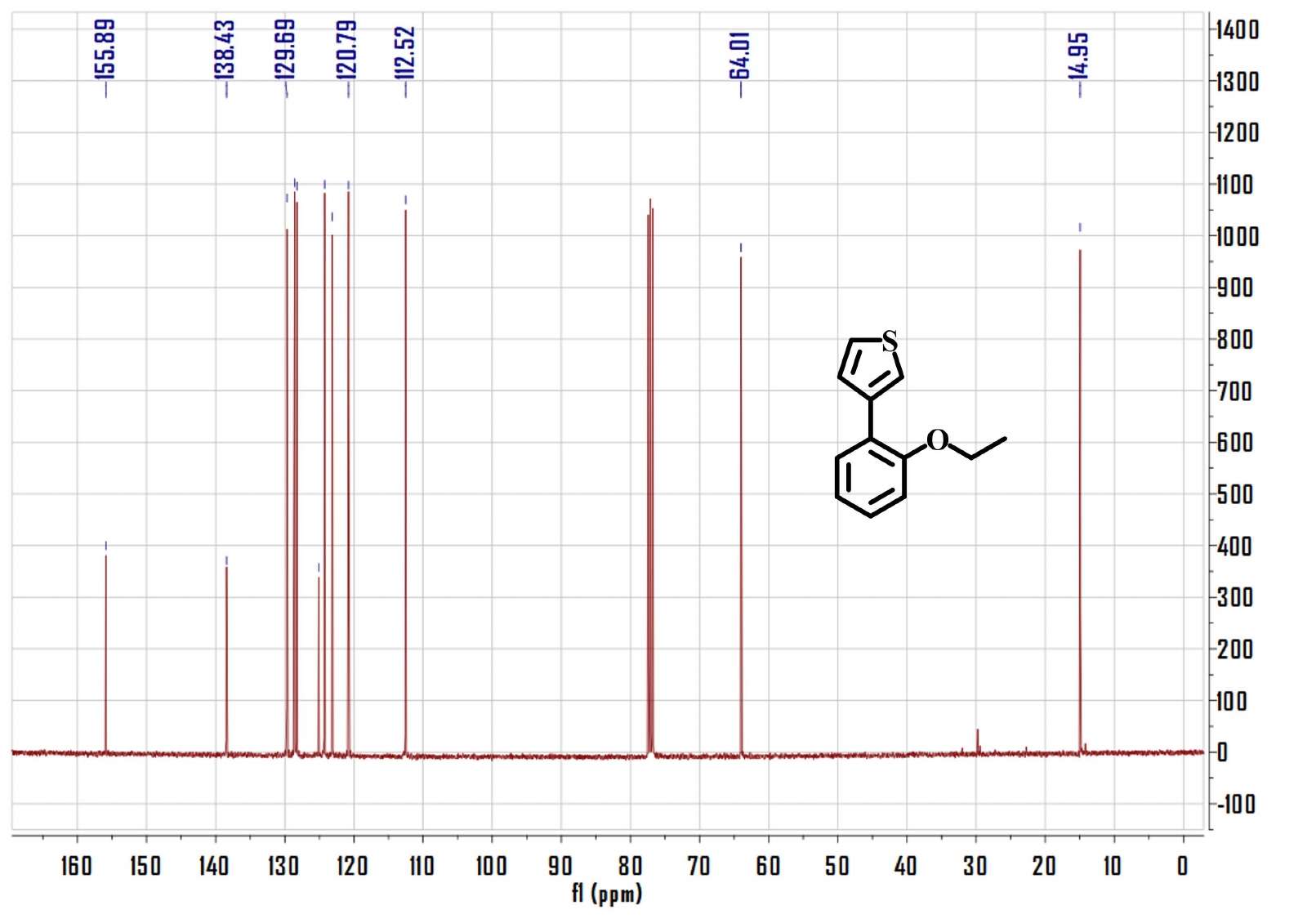}\hfill
    \includegraphics[width=0.32\linewidth]{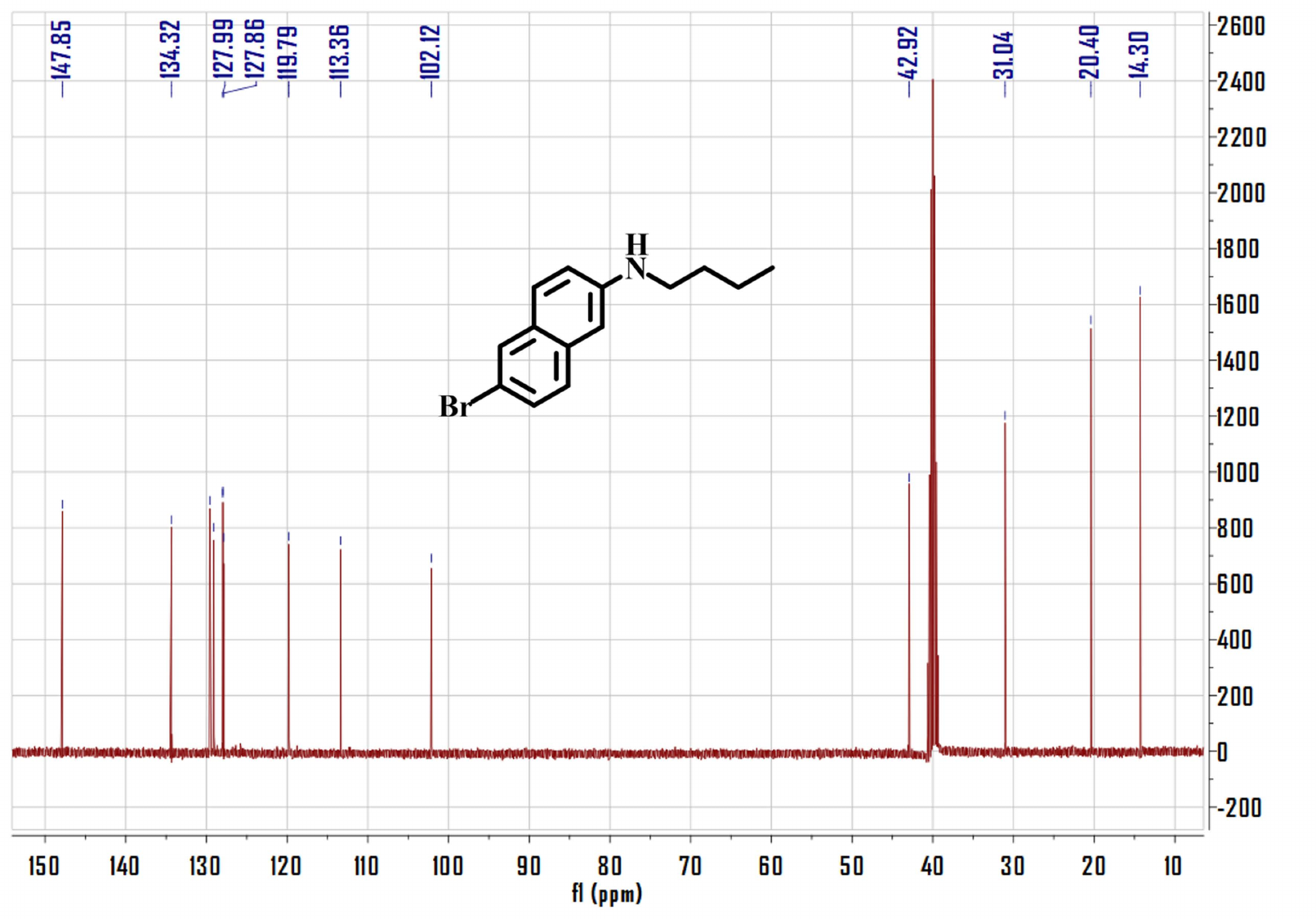}\hfill
    \includegraphics[width=0.32\linewidth]{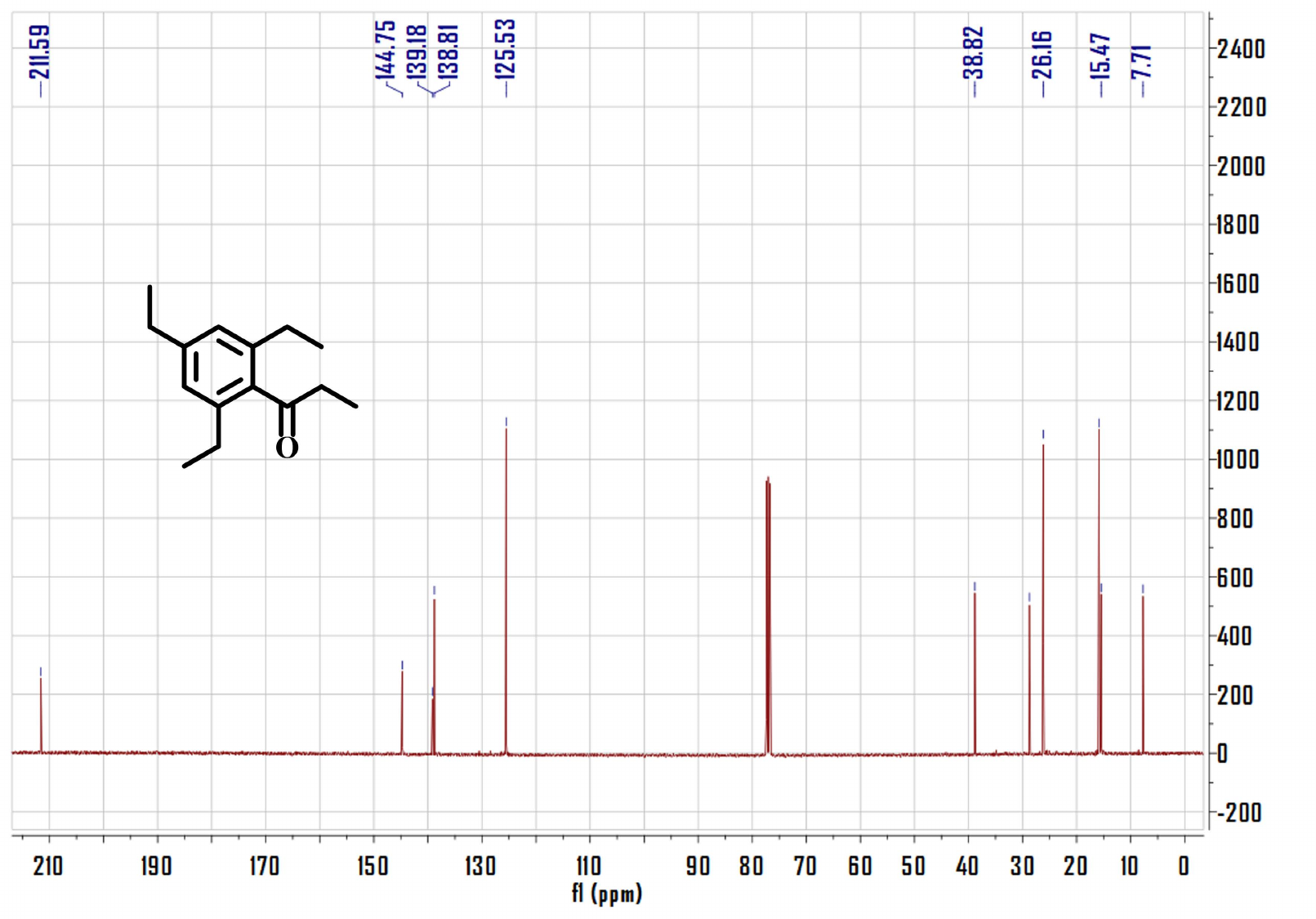}
    \caption{¹³C NMR spectra of all chemical products. \label{fig:Cpu}}
    \label{fig:all_spectra} 
\end{figure}

To verify the generalization ability of our prediction model in real-world scenarios, we conducted experiments in a chemistry laboratory based on the model’s predictions. In chemistry, the identification of a chemical substance is typically achieved through the analysis of its spectra. The products were sent to a third-party testing institution for analysis, where nuclear magnetic resonance (NMR) spectroscopy was used to determine whether the reaction products matched the predictions.As shown in Fig.~\ref{fig:Hpu} and Fig.~\ref{fig:Cpu}, the following images show the hydrogen (¹H NMR) and carbon (¹³C NMR) spectra of the predicted reaction product, which were found to be in complete agreement with the model’s predictions after analysis. The order from left to right is Suzuki--Miyaura coupling, Bucherer reaction, and Friedel--Crafts acylation. 

\newpage
\section{S3. Inference algorithm with Dynamic Molecular Dictionary}

Algorithm~\ref{alg:infer} details the streamlined RetroMPA inference workflow, in which the update of the molecular dictionary $\mathcal{D}$ is optional. For a reaction with two reactants, given a product SMILES $P$, the method first retrieves the Top-1 reactant pair $(b_1, b_2)$ from a base retrosynthesis model. If the update is enabled, the molecular embeddings of these base predictions are generated via Mol-Former and used to update $\mathcal{D}$. The refinement stage employs RetroMPA (denoted as $\text{Imp}$) to predict the remaining reactant conditioned on the product and a single reactant drawn from the base pair. Specifically, one reactant (e.g., $b_1$) is taken as the fixed condition, and $\text{Imp}$ generates a Top-$K$ candidate list $\mathcal{C}$ for the other reactant. A consistency check then compares the original counterpart $b_2$ against $\mathcal{C}$: if $b_2$ ranks within the top $K$, the base pair $(b_1, b_2)$ is retained; otherwise, the framework replaces $b_2$ with the highest-ranked candidate $\mathcal{C}[1]$.

\begin{algorithm}[htbp]
  \captionsetup{position=top}
  \setlength{\abovecaptionskip}{-0.5em}
  \caption{Streamlined RetroMPA Inference with Optional Dynamic Dictionary}
  \label{alg:infer}
  \begin{algorithmic}[1]
    \Require Product SMILES $P$, Base Model, Refinement Model $\text{Imp}$, Dynamic Dictionary $\mathcal{D}$ (optional), Threshold $K=10$, Boolean flag $\mathit{update\_dict}$
    \Ensure Final reactant combination $R_{\text{final}}$, (optionally) updated dictionary $\mathcal{D}$
    
    \State $(b_1, b_2) \leftarrow \text{BaseModel.Top1}(P)$ \Comment{Obtain initial Top-1 pair}
    
    \If{$\mathit{update\_dict}$}
      \For{each reactant $r \in \{b_1, b_2\}$}
        \State $\mathcal{D} \leftarrow \mathcal{D} \cup \{ (\text{SMILES}(r),\ \text{Mol-Former}(r)) \}$ \Comment{Optionally update dictionary}
      \EndFor
    \EndIf
    
    \State $r_{\mathrm{cond}} \leftarrow \text{RandomChoice}(\{b_1, b_2\})$ \Comment{Select one reactant as condition}
    \State $r_{\mathrm{target}} \leftarrow \{b_1, b_2\} \setminus \{r_{\mathrm{cond}}\}$ \Comment{The reactant to be predicted}
    \State $\mathcal{C} \leftarrow \text{Imp}(P, r_{\mathrm{cond}})$ \Comment{Generate Top-$K$ candidates for $r_{\mathrm{target}}$}
    
    \If{$r_{\mathrm{target}} \in \mathcal{C}[1:K]$}
      \State $R_{\text{final}} \leftarrow (b_1, b_2)$ \Comment{Keep original base prediction}
    \Else
      \State $R_{\text{final}} \leftarrow (r_{\mathrm{cond}}, \mathcal{C}[1])$ \Comment{Override with highest-ranked refined candidate}
    \EndIf
    
    \State \Return $R_{\text{final}}$, $\mathcal{D}$
  \end{algorithmic}
\end{algorithm}